\documentclass[journal,comsoc]{IEEEtran}

\usepackage[T1]{fontenc}

\usepackage{times}
\usepackage{epsfig}
\usepackage{graphicx}
\usepackage{amsmath}
\usepackage{amssymb}

\usepackage{algorithm}  
\usepackage{algpseudocode}
\usepackage{xcolor}
\usepackage{pifont}
\usepackage{multirow}
\usepackage{bbm}
\usepackage{mathtools}

\usepackage{subfigure}

\usepackage[pagebackref=true,breaklinks=true,letterpaper=true,colorlinks,bookmarks=false]{hyperref}
\usepackage[normalem]{ulem}
\begin{document}

\title{Animatable Neural Radiance Fields from Monocular RGB Videos}

\author{Jianchuan~Chen,
        Ying~Zhang,
        Di~Kang,
        Xuefei~Zhe,
        Linchao~Bao,
        Xu~Jia,
        Huchuan~Lu,~\IEEEmembership{Senior~Member,~IEEE}}

\maketitle

\begin{abstract}
We present animatable neural radiance fields (\textit{animatable NeRF}) for detailed human avatar creation from monocular videos. Our approach extends neural radiance fields (NeRF) to the dynamic scenes with human movements via introducing explicit pose-guided deformation while learning the scene representation network. In particular, we estimate the human pose for each frame and learn a constant canonical space
for the detailed human template, which enables natural shape deformation from the observation space to the canonical space under the explicit control of the pose parameters. 
To compensate for inaccurate pose estimation, we introduce the pose refinement strategy that updates the initial pose during the learning process, which not only helps to learn more accurate human reconstruction but also accelerates the convergence. In experiments we show that the proposed approach achieves 1) implicit human geometry and appearance reconstruction with high-quality details, 2) photo-realistic rendering of the human from novel views, and 3) animation of the human with novel poses.
\end{abstract}

\begin{IEEEkeywords}
Novel view synthesis, 3D human reconstruction, Image generation, Animation, Neural Rendering, Reconstruction algorithms. 
\end{IEEEkeywords}

\IEEEpeerreviewmaketitle

\section{Introduction}

\IEEEPARstart{T}{he} 3D human digitization has a wide range of applications in industries such as film, animation, games, and virtual try-on. Existing approaches to obtain high-quality 3D human reconstruction often require expensive equipment such as multiple synchronized cameras~\cite{textured_neural_avatars}, and RGB-D sensor~\cite{selfportraits}, limiting their applications in practical scenarios.  On the other hand, for various 3D human reconstruction approaches~\cite{PIFu,PIFuHD,Detailed_Human_Avatars,Video_avatars,MonoClothCap}, modeling complex geometric details such as hair, glasses, and cloth wrinkles of real humans remains a challenging problem.

\par
In this paper, we target to obtain photo-realistic 3D human avatars from monocular RGB videos, which are one of the most accessible video forms in daily life. Different from existing approaches based on pre-scanned human models~\cite{AMAS} or parametric body models~\cite{Detailed_Human_Avatars, MonoClothCap, Video_avatars}, our approach implicitly reconstructs the human geometry and appearance via generalizing Neural Radiance Fields (NeRF)~\cite{NeRF}, which uses a neural network to encode color and density as a function of location and viewing angle, and generates photo-realistic images by volume rendering. 
NeRF~\cite{NeRF} has shown impressive ability in reconstructing a static scene from multi-view images, and inspires many researchers in extending NeRF to scenes with severe lighting changes~\cite{nerf-w} and non-rigid deformations~\cite{nerfies,NGNeRF}. However, these approaches are uncontrollable and limited to nearly static scenes with small movements, failing to deal with human subjects with large movements.

To handle the dynamic human from monocular videos, we combine neural radiance fields with a parametric body model of SMPL~\cite{SMPL}, which enables more precise human geometry and appearance modeling, and further makes the neural radiance fields controllable. 
In particular, our approach extends NeRF via introducing the pose-guided deformation, which unwarps the observation space near the human body to a constant canonical space through the deformation of the SMPL.
We observe that even the state-of-the-art SMPL estimator from monocular videos cannot obtain accurate parameters, which inevitably leads to blurry results. To address this problem, we propose to jointly optimize the NeRF and SMPL parameters via analysis-by-synthesis, which not only obtains better results but also accelerates the convergence of training. 
We demonstrate the superiority of the proposed method on multiple datasets, with both quantitative and qualitative results on novel view synthesis, 3D human reconstruction, and novel pose synthesis.

In summary, our work has the following contributions:
\begin{itemize}
    \item We propose a method explicitly deforming the points according to SMPL pose to reconstruct a canonical view NeRF model, relaxing the requirement of the static object and preserving details such as clothing and hair.
    \item We incorporate pose refinement into our analysis-by-synthesis approach to account for the inaccurate SMPL estimates, resulting in refined SMPL pose and greatly improved reconstruction quality.
    \item We achieve high-quality 3D human reconstruction from monocular RGB video, and can render photo-realistic images from novel views.
    \item Due to our controllable SMPL-based geometry deformation, we can synthesize novel pose images, showing that our learned canonical space NeRF model is animatable.
\end{itemize}


\section{Related work}

\textbf{3D Human Reconstruction}. Reconstructing 3D human has been more and more popular in recent years. 
Various approaches attempt to digitize a human from a single-view image~\cite{PIFu,PIFuHD,360degree,tex2shape,ARCH}, multi-view images~\cite{PIFu,PIFuHD,dvv,PaMIR}, RGB videos~\cite{Video_avatars, Detailed_Human_Avatars,lrp,MonoClothCap,DoubleFusion}, or RGB-D videos~\cite{selfportraits,texmesh}. One stream of these approaches~\cite{Detailed_Human_Avatars, Video_avatars, MonoClothCap} utilize a parametric body model such as SMPL~\cite{SMPL} to represent a human body with cloth deformations, which produces an animatable 3D model with high-quality textures but struggles with limited expressive ability in complex geometries such as hair and dresses. On the other hand, PIFu~\cite{PIFu} and PIFuHD~\cite{PIFuHD} based methods use an implicit representation to reconstruct a 3D surface and achieves impressive results in handling people with complex poses, hairstyles and clothing, which however, suffers from a blurry appearance and requires further registration for animation. To handle more complex pose inputs, these methods~\cite{ARCH,PaMIR,S3,ipnet} combine implicit representations and parametric models to obtain more robust results and are animatable.
    
\textbf{Neural Representations}. Representing a scene with neural networks has achieved stunning success in recent years.
SRN~\cite{SRN} proposes an implicit neural representation that assigns feature vectors to 3D positions, and uses a differentiable ray marching algorithm for image generation.
NeRF~\cite{NeRF} establishes a static scene that maps 3D coordinates and viewing direction to density and color using a neural network. These methods~\cite{SRN,NeRF,NSVF} can render very realistic images, but they are all limited to static scenes.
Dynamic NeRFs~\cite{nerfies,D-NeRF,NSVF,NGNeRF,NSFF,HyperNeRF} extend NeRF to dynamic scenes by introducing the latent deformation field or scene flow fields.
These methods where the deformations are learned by networks allow to handle more general deformation, and synthesize novel poses by using interpolation in the latent space. 
However, it is difficult to implicitly control the complex non-rigid deformation of human body motion.
These works~\cite{NerFACE,Neural_Body,A-NeRF,SCANimate,SMPLicit,LoopReg} combine scene representation network with parametric models~\cite{3DMM,SMPL} to reconstruct dynamic humans.
Instead of using latent codes or expression parameters as input, we use the human body model SMPL to explicitly deform over different poses and shapes and reconstruct the human body in the canonical pose.
At the same time, this explicit method allows us to fine-tune the parameters of the SMPL which is very practical in real scenarios.
Similar ideas with us have been used in recent works\cite{Neural_Body, Anim-NeRF, Neural_Actor}, but these methods usually require multi-view images or accurate registered SMPL.
It is more challenging for monocular videos because of the difficulty of SMPL estimation.

\textbf{Human Motion Transfer}.
Human motion transfer aims to synthesize an image of a person with the appearance from a source human and the motion from a reference image. Recent advances using Generative Adversarial Networks (GAN) have shown convincing performance without recovering detailed 3D geometry.
These works~\cite{monkey_net,eveybody_dance_now,LWGAN,LWGANPlusPlus} use image-to-image translation~\cite{pix2pix,pix2pixhd} to map 2D skeleton images to rendering output. Due to the lack of 3D reasoning, the geometry of the generated humans is usually not consistent across multiple views and motions.
To better preserve the appearance of the source subject, these methods ~\cite{StylePoseGAN,HumanGAN} use UV map to transform features from screen space to UV texture space to obtain the neural texture, then render the feature maps in the target pose by neural rendering network.
In addition to these general approaches between arbitrary subjects, there are other person-specific methods~\cite{textured_neural_avatars,NHR,smplpix}.
Textured Neural Avatar~\cite{textured_neural_avatars} learns a uniform neural texture from different views and poses.
SMPLpix\cite{smplpix} and NHR\cite{NHR} project the point clouds to 2D images and then feed them into an image-to-image translation network.
However, these neural rendering methods fail to generate photorealistic results for novel poses that were not seen during training.


\begin{figure*}[ht]
    \begin{center}
	\begin{tabular}{@{}c}
		\includegraphics[width=0.8\linewidth, ]{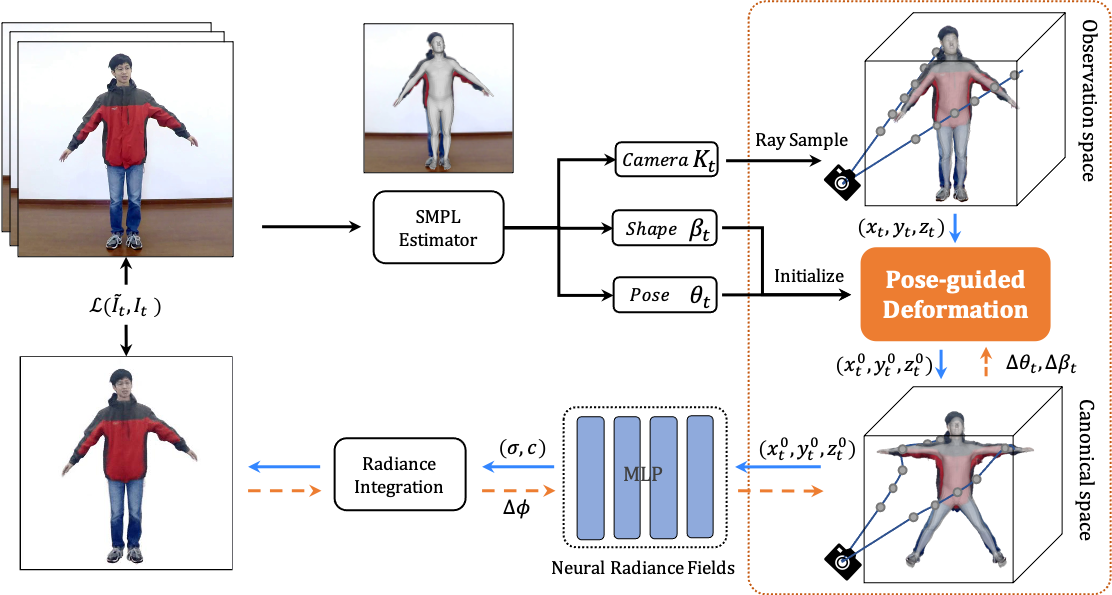} 
	\end{tabular}
	\end{center}
	\caption{Overview of the proposed Animatable Neural Radiance Fields. Given a video sequence, we estimate the camera $K_t$ and SMPL parameters $M(\theta_{t}, \beta_{t})$ of the human subject for initialization. We use volume rendering to sample points $(x_t, y_t, z_t)$ along the camera ray in observation space, and transform these points to canonical space according to pose-guided deformation. Then we input these points $(x_t^0, y_t^0, z_t^0)$ into the neural radiance field to get densities $\sigma$ and colors $\mathbf{c}$. Then we use the integral equation Eq.~\eqref{eq:rendering equation} to render the image, and jointly optimize the neural radiance field parameters $\phi$ and SMPL parameters  $\theta_t,\beta_t$ by minimizing the error $\mathcal{L}\left(\tilde{I}_t, I_t\right)$ between the rendered image $\tilde{I}_t$ and the ground truth image $I_t$ with the mask.} \label{fig:framework}
\end{figure*}

\section{Method}
In this section, we will describe the method to create a human avatar from a single portrait video of a person as shown in Fig.~\ref{fig:framework}.
Given a $n$-frame video sequence  $\left\{I_t\right\}_{t=1}^{n}$ of a single human subject turning around before the camera and holding an A-pose or T-pose, we estimate the SMPL~\cite{SMPL,SMPL-X} parameters $M(\theta_{t}, \beta_{t})$ and camera intrinsics $K_{t}$ of each frame using existing human body shape and pose estimation models~\cite{VIBE}. 
In order to avoid the influence of background changes caused by camera movement, we first use a segmentation network~\cite{RP-R-CNN} to obtain the foreground human mask and set the background color to white uniformly.
Our animatable NeRF (Sec.~\ref{Anim-NeRF}) can be decomposed into pose-guided deformation (Sec.~\ref{Pose-Guided_deform}) and a neural radiance field (NeRF) defined in the canonical space. We can use the volume rendering (Sec.~\ref{Volume_Render}) to render our neural radiance field. In order to avoid the negative effects of inaccurate SMPL parameters, we propose to jointly optimize the neural radiance field and SMPL parameters (Sec.~\ref{Pose_Refine}). We also introduce background regularization and pose regularization to improve the robustness of optimization (Sec.~\ref{Object_Fun}).

\subsection{Animatable Neural Radiance Fields}\label{Anim-NeRF}
To model human appearance and geometry with complex non-rigid deformation, we introduce the parameterized human model SMPL~\cite{SMPL} into the neural radiance field and present the animatable neural radiance fields (animatable NeRF) $F$ which maps the 3D position $\mathbf{x}=(x, y, z)$, shape $\mathbf{\beta}_{t}$ and pose $\mathbf{\theta}_{t}$ into color $\mathbf{c}=(r, g, b)$ and density $\sigma$:
\begin{equation}
F\left(D\left(\mathbf{x}, \mathbf{\theta}_{t}, \mathbf{\beta}_{t} \right)\right) = \left(\mathbf{c},\sigma\right)
\end{equation}
where $D\left(\mathbf{x}, \mathbf{\theta}_{t}, \mathbf{\beta}_{t} \right)$ transforms the 3D position $\mathbf{x}=(x, y, z)$ in the observation space to $\mathbf{x}^{0}=(x^{0}, y^{0}, z^{0})$ in canonical space, aiming to handle human movements between different frames. The view dependence in NeRF is mainly for dealing with specular reflections of materials such as metal and glass. But the skin and clothes of humans are mainly diffuse reflective materials, so we remove the viewing direction from the input.  We will discuss the impact of viewing direction in Sec.~\ref{impact_of_view_direction}.

\subsection{Pose-guided Deformation}\label{Pose-Guided_deform}
In contrast to~\cite{NGNeRF,nerfies} that implicitly control of the deformation of spatial points, we use the parametric body model - SMPL, to explicitly guide the deformation of spatial points. 
%
Here we define the observed image as the observation space and attempt to learn a template human in the canonical space.
The articulated SMPL model enables the explicit transformation of the spatial points (i.e. from observation space to canonical space), which facilitates the learning of a specified meaningful canonical space, and reduces the reliance of diverse input poses to generalize to unseen poses so that we can learn the NeRF space from dynamic scenes (containing moving person) and animate this person after training.
The template pose in the canonical space is defined as X-pose $\theta^{0}$ (as shown in Fig.~\ref{fig:framework}), due to its good visibility and separability of each body part.
By using the inverse transformation of the linear skinning of SMPL, the pose $\theta_{t}$ in observation space can be transformed into the X-pose $\theta^{0}$ in canonical space. Considering that the transformation functions are only defined on the surface vertices of the body mesh, we extend them to the space near the mesh surface based on the intuition that points in space near the mesh should move along with neighboring vertices. Following PaMIR~\cite{PaMIR} we define the transformation of a point $\mathbf{x}$ from observation space to canonical space as

\begin{equation}
\label{eq:transformation}
\begin{aligned}
\left[\begin{matrix}\mathbf{x}^{0} \\ 1\end{matrix} \right]&=\mathbf{M}(\mathbf{x}, \beta_{t}, \theta_{t}, \theta^{0}) \left[\begin{matrix}\mathbf{x} \\ 1\end{matrix} \right] \\
\mathbf{M}(\mathbf{x}, \beta_{t}, \theta_{t}, \theta^{0}) &=\sum_{v_i \in \mathcal{N}\left(\mathbf{x}\right)} \frac{\omega_{i}}{\omega} \mathbf{M}_{i}\left(\beta_{t}, \theta^{0}\right)\left(\mathbf{M}_{i}(\beta_{t}, \theta_{t})\right)^{-1}
\end{aligned}
\end{equation}
where $\mathcal{N}\left(\mathbf{x}\right)$ denotes the SMPL vertex set near $\mathbf{x}$ in the observation space, and Eq.~\eqref{eq:transformation} indicates that the movement of $\mathbf{x}$ relies on the movement of neighboring vertices. 
The transformation weight $\omega_{i}$ that the vertex $v_i$ affects the point $\mathbf{x}$ is defined as 

\begin{equation}
\begin{aligned}
\omega_{i} &=\exp \left(-\frac{\left\|\mathbf{x}-v_{i}\right\| \left\|\hat{\mathbf{b}}-\mathbf{b}_{i}\right\|}{2 \sigma^{2}}\right)  \\
\omega &=\sum_{v_i \in \mathcal{N}(\mathbf{x})} \omega_{i}
\end{aligned}
\end{equation}
where $\mathbf{b}_{i}$ is the blend weight of $v_i$ and $\hat{\mathbf{b}}$ is the blend skinning weight of the nearest vertex, and $\|\mathbf{x}-v_{i}\|$ computes the L2 distance between $\mathbf{x}$ and $v_i$.
Consider the fact that a point might be affected by different body parts, leading to ambiguous or even non-meaningful transformation,  we adopt the blend skinning weight which characterizes the movement patterns of a vertex along with the SMPL joints~\cite{SMPL}, to strengthen the movement impact of the nearest neighbor. $\omega$ is used for weight normalization.

Following SMPL\cite{SMPL}, the transformation matrix $\mathbf{M}_{i}\left(\beta, \theta\right)$ of mesh vertex $v_i$ from rest pose to $\theta$-pose is computed by
\begin{equation}
\mathbf{M}_{i}(\beta, \theta) = \left(\sum_{j=1}^{K}b_{i,j}\mathbf{G}_j \right)
\begin{bmatrix}
  \mathbf{I}  & \mathbf{B}_{S,i}(\beta)+\mathbf{B}_{P, i}(\theta) \\
    \mathbf{0}^T & 1
\end{bmatrix}
\end{equation}
where $\mathbf{G}_j\in \mathbb{R}^{4\times4}$ is the world transformation of joint $j$, $b_{i,j}$ is the blend skinning weight representing how much the rotation of part $j$ affects vertex $v_i$, $\mathbf{B}_{P, i}(\theta)\in \mathbb{R}^{3}$ and $\mathbf{B}_{S,i}(\beta)\in \mathbb{R}^{3}$ are the pose blendshape and shape blendshape of vertex $v_i$ respectively. 

\subsection{Volume Rendering}\label{Volume_Render}
\label{sec:vr}
We use the same volume rendering techniques as in NeRF~\cite{NeRF} to render the neural radiance field into a 2D image. For a given video frame $I_t$, we first convert the camera coordinate system to the SMPL coordinate system, that is, transform the SMPL global rotation and translation to the camera. Then the pixel colors are obtained by accumulating the colors and densities along the corresponding camera ray $\mathbf{r}$. In practice, the continuous integration is approximated by sampling $N$ points $\left\{\mathbf{x}_{k}\right\}_{k=1}^{N}$ between the near plane and the far plane along the camera ray $\mathbf{r}$ as
\begin{equation}
\label{eq:rendering equation}
\begin{array}{c}
\tilde{C}_{t}(\mathbf{r})=\sum\limits_{k=1}^{N}T_{k}\left(1-\exp \left(\eta_{t}(\mathbf{x}_{k})\sigma_{t}\left(\mathbf{x}_{k}\right) \delta_{k}\right)\right) \mathbf{c}_{t}\left(\mathbf{x}_{k}\right) \\
\tilde{D}_{t}(\mathbf{r})=\sum\limits_{k=1}^{N}T_{k}\left(1-\exp \left(\eta_{t}(\mathbf{x}_{k})\sigma_{t}\left(\mathbf{x}_{k}\right) \delta_{k}\right)\right) \\
T_{k}=\exp \left(-\sum\limits_{j=1}^{k-1} \eta_{t}(\mathbf{x}_{k})\sigma_{t}\left(\mathbf{x}_{j}\right) \delta_{j}\right)
\end{array}
\end{equation}
where $\delta_{k}=\left\|\mathbf{x}_{k+1}-\mathbf{x}_{k}\right\|_{2}$ is the distance between adjacent sampling points, and $\eta_{t}(\mathbf{x}_{k})$ is a prior 3D \textit{mask} (detailed in the following) used to provide geometric prior guidance and deal with ambiguity during pose-guided deformation.

Since we only focus on modeling a single human subject, here we introduce an assumption for learning a more accurate neural radiance field: The densities should be zeros for points far from the surface of human mesh;
\begin{equation}
\label{distance constraint}
\begin{aligned}
\eta_{t}(\mathbf{x}_{k}) &=  d(\mathbf{x}_{k}) \leq \delta\\
d(\mathbf{x}_{k}) &= \sum_{v_i \in \mathcal{N}\left(\mathbf{x}_{k}\right)} \frac{\omega_{j}}{\omega} \left\|\mathbf{x}_{k}-v_{i}\right\|
\end{aligned}
\end{equation}
where $d(\mathbf{x}_{k})$ is the weighted distance from point $\mathbf{x_k}$ to the nearest neighbor vertices $\mathcal{N}\left(\mathbf{x}_{k}\right)$ in the observation space. $\delta$ is the distance threshold limiting distance between the sample point to the SMPL surface in the observation space. In experiments we follow NeRF~\cite{NeRF} to perform hierarchical volume sampling to obtain $\tilde{C}_{t}^c(\mathbf{r})$ and $\tilde{C}_{t}^f(\mathbf{r})$ with the coarse and fine networks, respectively.

\subsection{Pose Refinement via Analysis-by-Synthesis} \label{Pose_Refine}
\begin{figure*}[t]
    \centering
    \subfigure[NeRF]{
	\begin{minipage}[b]{0.14\linewidth}
	\includegraphics[width=1.0\linewidth, trim=0 0 0 0,clip]{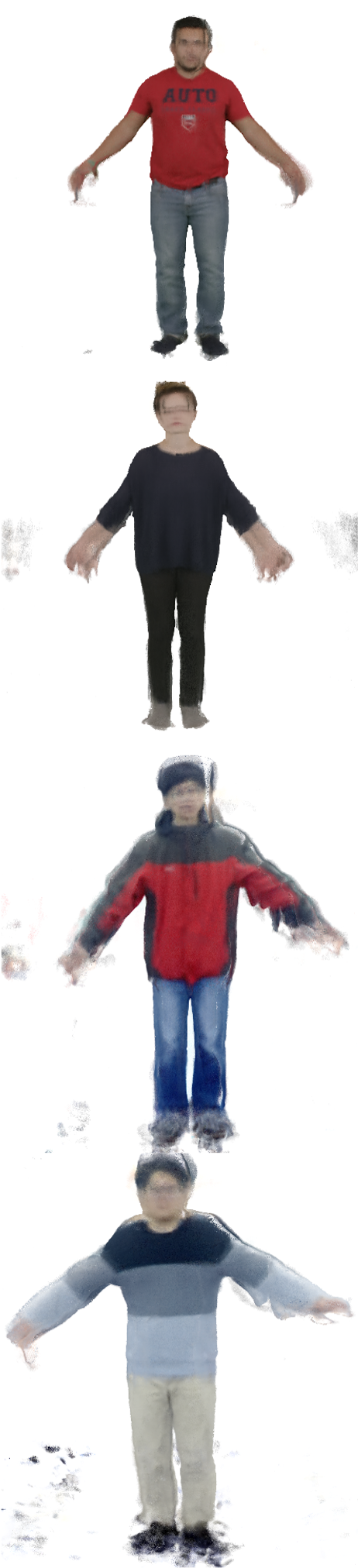}
	\end{minipage}
	}
	\subfigure[SMPLpix]{
	\begin{minipage}[b]{0.14\linewidth}
	\includegraphics[width=1.0\linewidth, trim=0 0 0 0,clip]{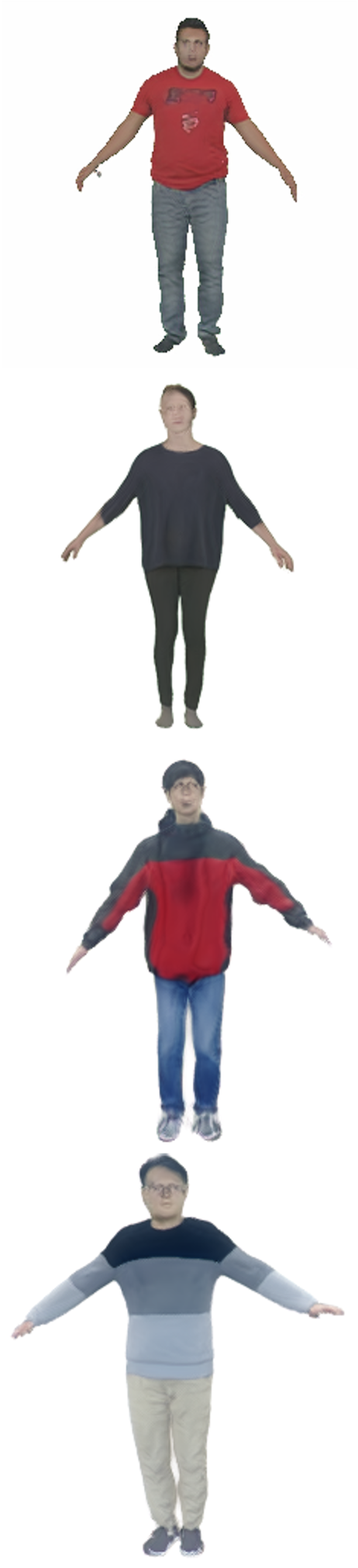}
	\end{minipage}
	}
	\subfigure[NeuralBody]{
	\begin{minipage}[b]{0.14\linewidth}
	\includegraphics[width=1.0\linewidth, trim=0 0 0 0,clip]{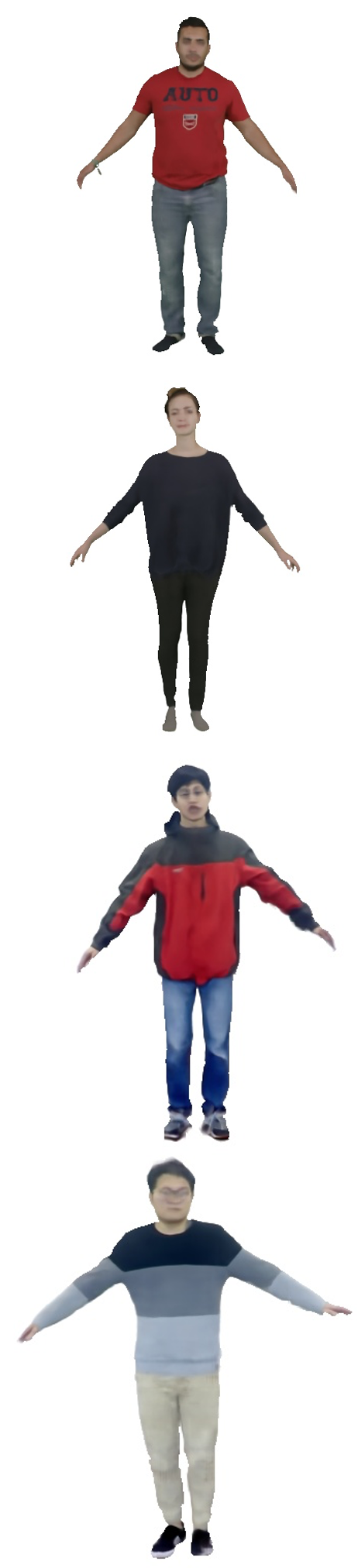}
	\end{minipage}
	}
	\subfigure[NeRF+U]{
	\begin{minipage}[b]{0.14\linewidth}
	\includegraphics[width=1.0\linewidth, trim=0 0 0 0,clip]{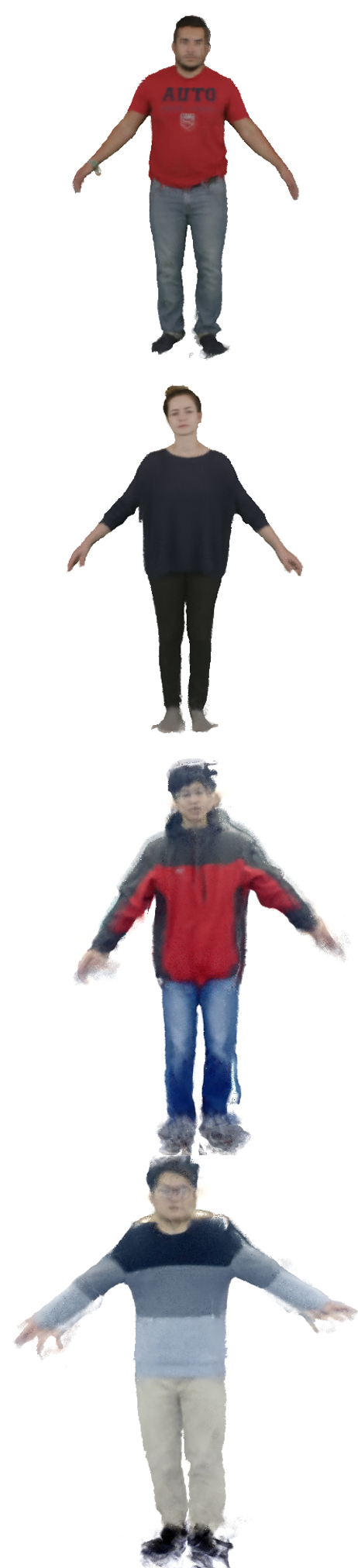}
	\end{minipage}
	}
	\subfigure[Ours]{
	\begin{minipage}[b]{0.14\linewidth}
	\includegraphics[width=1.0\linewidth, trim=0 0 0 0,clip]{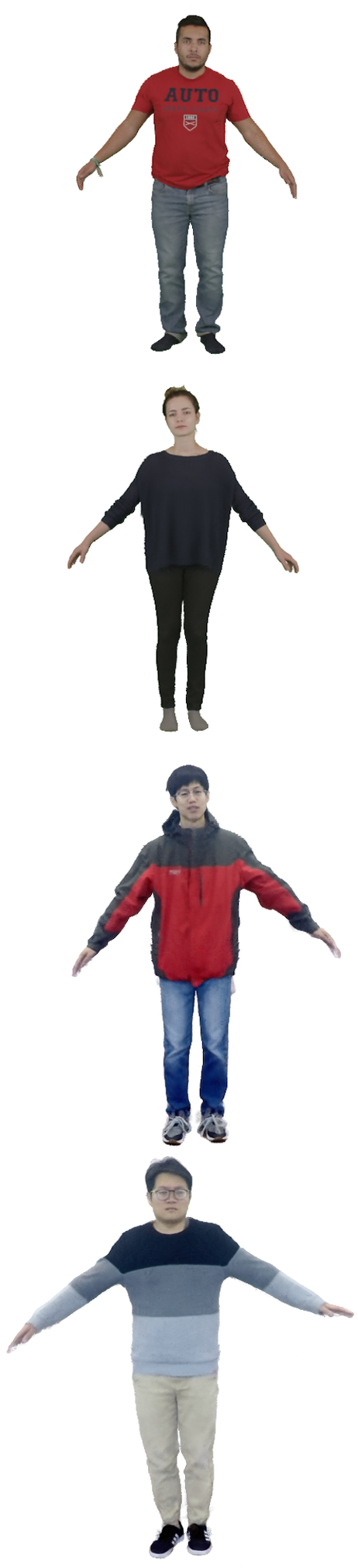}
	\end{minipage}
	}
	\subfigure[GT]{
	\begin{minipage}[b]{0.14\linewidth}
	\includegraphics[width=1.0\linewidth, trim=0 0 0 0,clip]{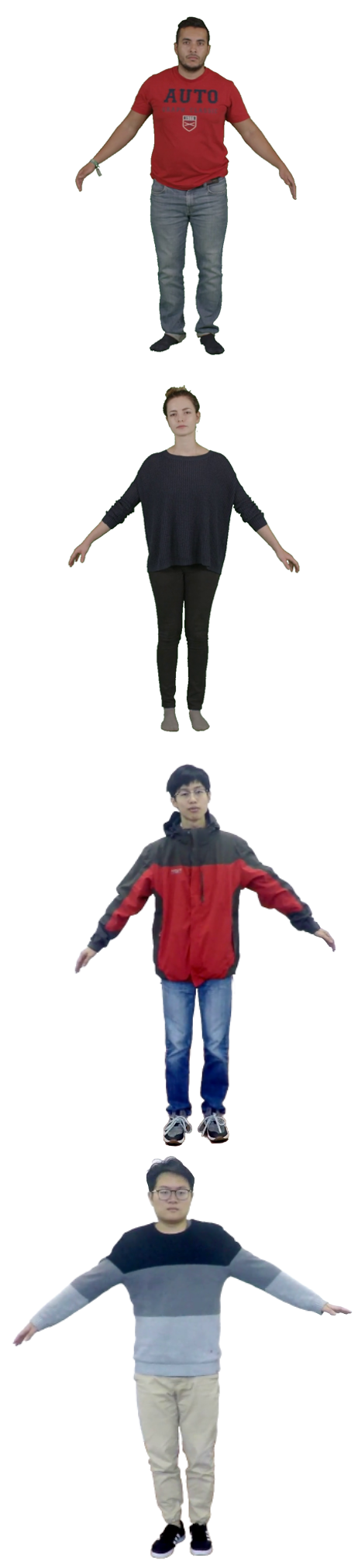}
	\end{minipage}
	}
    \caption{Visual comparison of different methods about novel view synthesis on People-snapshot\cite{Video_avatars}(1-2 rows) and iPER\cite{LWGAN}(3-4 rows). 
    NeRF\cite{NeRF} is struggling to handle dynamic scenes because the movement of the subject violates the multi-view consistency requirement. With the help of our proposed pose-guide deformation, NeRF+U (NeRF + Unpose) achieves much better results (row 1\&2) if the estimated SMPL poses are accurate but still produces blurry results (row 3\&4) if they are not.
    Further adding pose refinement (ours) greatly improves the robustness as long as the estimated SMPL pose is reasonably good.
    Compared with NeuralBody\cite{Neural_Body} and SMPLpix\cite{smplpix}, our approach can produce realistic images with well preserved identity and cloth details.
    }
    \label{fig:Comparision_with_different_method}
\end{figure*}

Our proposed method learns an animatable NeRF for human subjects via explicitly deforming the observation space from different frames to a constant canonical space, under the guidance of SMPL transformations. Although the current state-of-the-art pose and shape estimation methods~\cite{VIBE} is adopted to obtain more stable SMPL parameters, in experiments we observe that results estimated by these methods do not align well with the ground truth, especially in depth. The inaccurate human body estimation could easily lead to blurry results. To address this problem, we propose to fine-tune the SMPL parameters during training. 
Specifically, we use VIBE~\cite{VIBE} to estimate SMPL parameters $M(\theta_{t}, \beta_{t})$ for each frame $I_t$ as initialization of variables, which will be optimized during training.
We use the mean shape parameters $\beta=\frac{1}{n}\sum_{t=1}^{n}\beta_{t}$ for different frames. It turns out that the refined SMPL can better fit the input image, and helps to obtain clearer and sharper results as shown in Fig.~\ref{fig:Comparision_with_different_method} and Fig.~\ref{fig:3D_Rec_on_mutil_garment}.


\subsection{Objective Functions} \label{Object_Fun}
Given a monocular video sequence, we learn the animatable NeRF by optimizing the following objective function
\begin{equation}
\mathcal{L}= \mathcal{L}_{c} + \mathcal{L}_{p} + \lambda_{d} * \mathcal{L}_{d}
\end{equation}
where $\mathcal{L}_{c}$, $\mathcal{L}_{p}$, and $\mathcal{L}_{d}$ are reconstruction loss, pose regularization, and background regularization, respectively. $\lambda_{d}$ aims to balance the importance of background regularization.

{\bf Reconstruction Loss}. The reconstruction loss aims to minimize the error between the rendered images and the corresponding observed frames, which is defined as 
\begin{equation}
\mathcal{L}_{c}=\sum_{t}\sum_{\mathbf{r}}\left\|\tilde{C}_{t}^{c}(\mathbf{r})-C_{t}(\mathbf{r})\right\|_{2}^{2} + \left\|\tilde{C}_{t}^{f}(\mathbf{r})-C_{t}(\mathbf{r})\right\|_{2}^{2}
\end{equation}
where $\mathbf{r}$ is a camera ray passing through the image $I_{t}$.
$C_t(\mathbf{r})$ is the ground truth color of the pixel intersected by ray $\mathbf{r}$ on the observed image $I_{t}$.
And $\tilde{C}_{t}^{c}(\mathbf{r})$, $\tilde{C}_{t}^{f}(\mathbf{r})$ are the corresponding rendered colors from the coarse and fine networks respectively (see Sec.~\ref{sec:vr}).

{\bf Pose Regularization}. 
To obtain stable and smooth pose parameters, we add the following pose regularization term to encourage the optimized pose parameter to stay close to the initial pose, and the pose parameters between adjacent frames to be similar.
\begin{equation}
\mathcal{L}_{p}=\lambda_1\left\|\tilde{\theta}_{t}-\theta_{t}\right\|+\lambda_2\left\|\tilde{\theta}_{t}-\tilde{\theta}_{t+1}\right\|
\end{equation}
where $\theta_{t}$ is the initial pose parameters, $\tilde{\theta}_{t}$ and $\tilde{\theta}_{t+1}$ are the optimized pose parameters of frame $t$ and $t+1$. $\lambda_1$ and $\lambda_2$ are the corresponding penalty weights.

{\bf Background Regularization}. 
We only focus on reconstructing the human no matter what the background is, which means, ideally, density exists only inside the human. 
To better achieve this goal, we first set the background color to white with the help of an off-the-shelf segmentation network.
We minimize the difference between the rendered integral density and the mask obtained by segmentation.
Since the foreground (i.e. human) region is $1$ and the background region is $0$ in the mask, we are encouraging the human region's integral density to be 1 and encouraging the background region's integral density to be 0, resulting in a much cleaner empty space estimation and more solid and clearer person estimation in our canonical NeRF space.
Mathematically, our background regularization term is defined as follows,
\begin{equation}
\mathcal{L}_{d}=\sum_{t}\sum_{\mathbf{r}}\left\|\tilde{D}_{t}^{c}(\mathbf{r})-D_{t}(\mathbf{r})\right\| + \left\|\tilde{D}_{t}^{f}(\mathbf{r})-D_{t}(\mathbf{r})\right\|
\end{equation}
where $\tilde{D}_{t}^{c}$ and $\tilde{D}_{t}^{f}$ is the rendered integral density of the coarse and fine network for the camera ray $\mathbf{r}$ from the image $I_{t}$, and $D_{t}(\mathbf{r})$ is the corresponding segmentation mask and $D_{t}(\mathbf{r})=1$ in the foreground region and $D_{t}(\mathbf{r})=0$ in the background region.

\begin{figure*}[ht]
    \centering
    \subfigure[Input]{
	\begin{minipage}[b]{0.1116\linewidth}
	\includegraphics[width=1.0\linewidth, trim=0 0 0 0,clip]{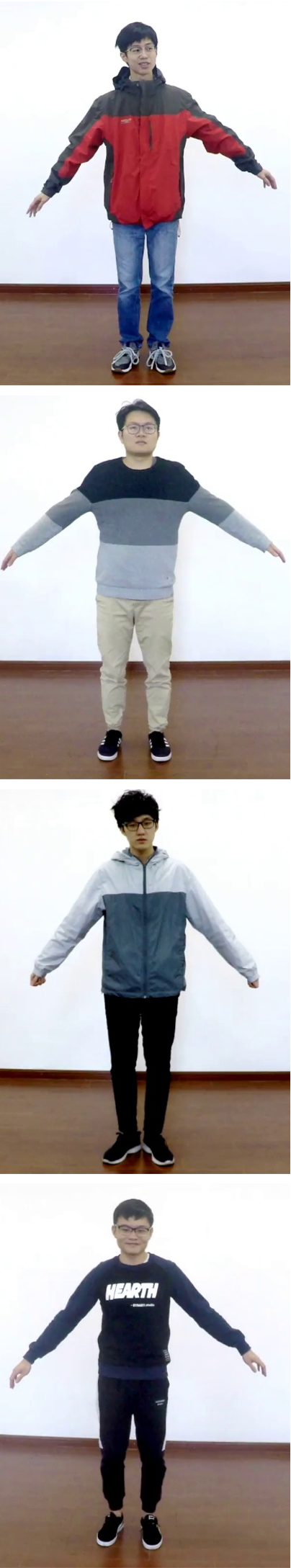}
	\end{minipage}
	}
	\subfigure[view 0]{
	\begin{minipage}[b]{0.1116\linewidth}
	\includegraphics[width=1.0\linewidth, trim=0 0 0 0,clip]{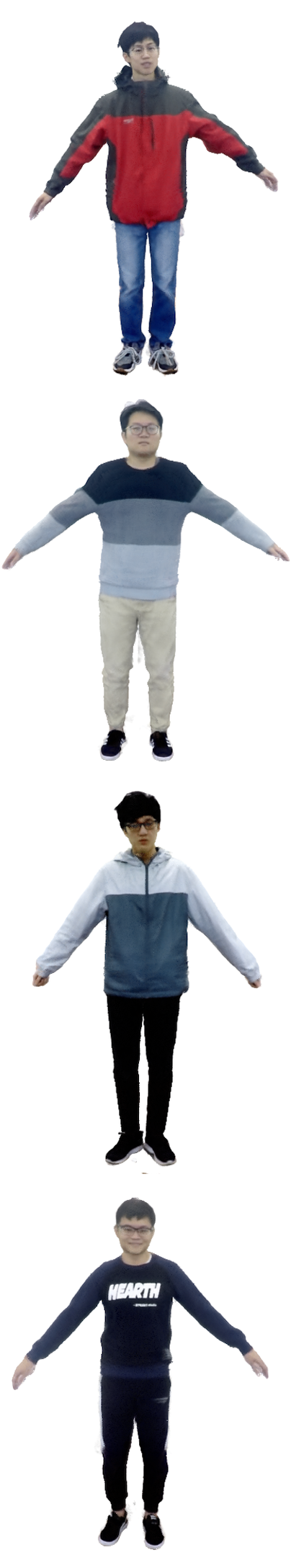}
	\end{minipage}
	}
	\subfigure[view 1]{
	\begin{minipage}[b]{0.10\linewidth}
	\includegraphics[width=1.0\linewidth, trim=0 0 0 0,clip]{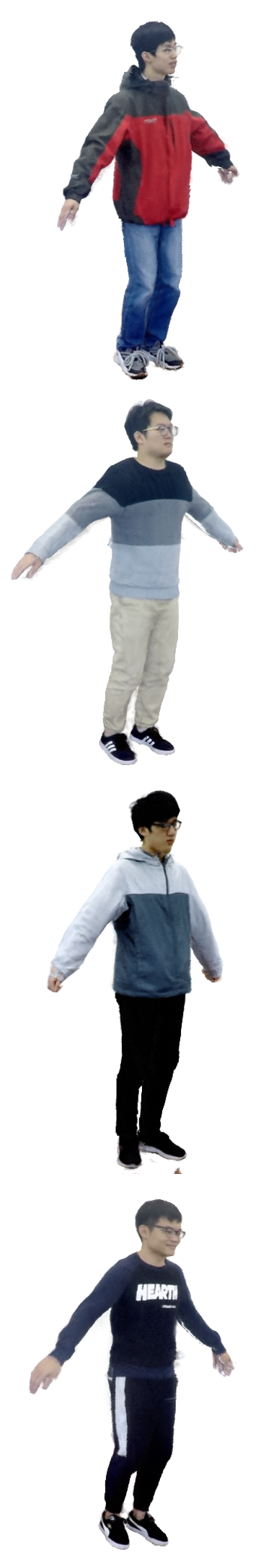}
	\end{minipage}
	}
	\subfigure[view 2]{
	\begin{minipage}[b]{0.10\linewidth}
	\includegraphics[width=1.0\linewidth, trim=0 0 0 0,clip]{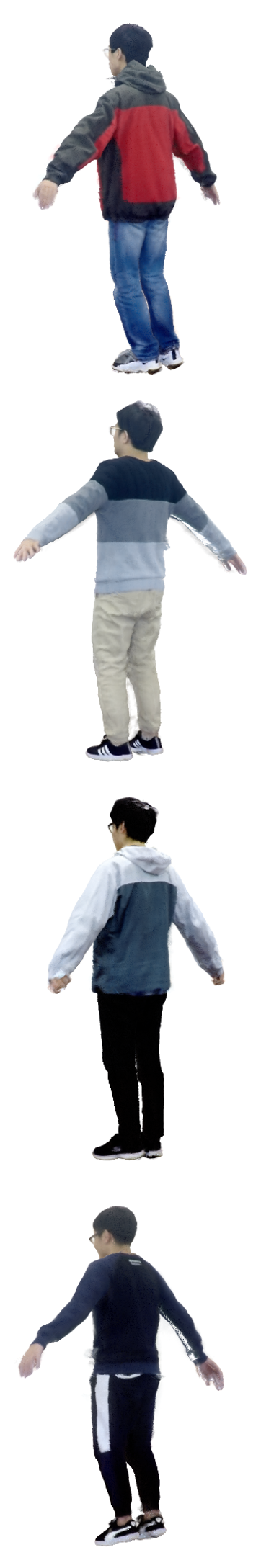}
	\end{minipage}
	}
	\subfigure[Input]{
	\begin{minipage}[b]{0.10\linewidth}
	\includegraphics[width=1.0\linewidth, trim=0 0 0 0,clip]{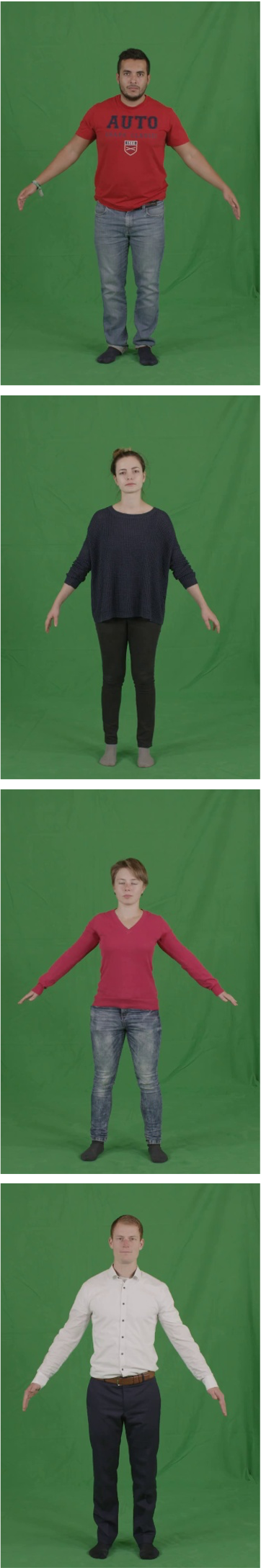}
	\end{minipage}
	}
	\subfigure[view 0]{
	\begin{minipage}[b]{0.10\linewidth}
	\includegraphics[width=1.0\linewidth, trim=0 0 0 0,clip]{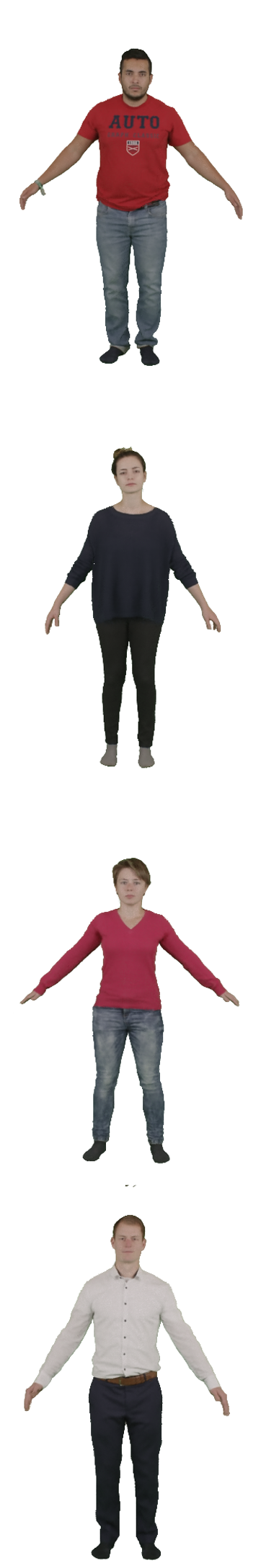}
	\end{minipage}
	}
	\subfigure[view 1]{
	\begin{minipage}[b]{0.10\linewidth}
	\includegraphics[width=1.0\linewidth, trim=0 0 0 0,clip]{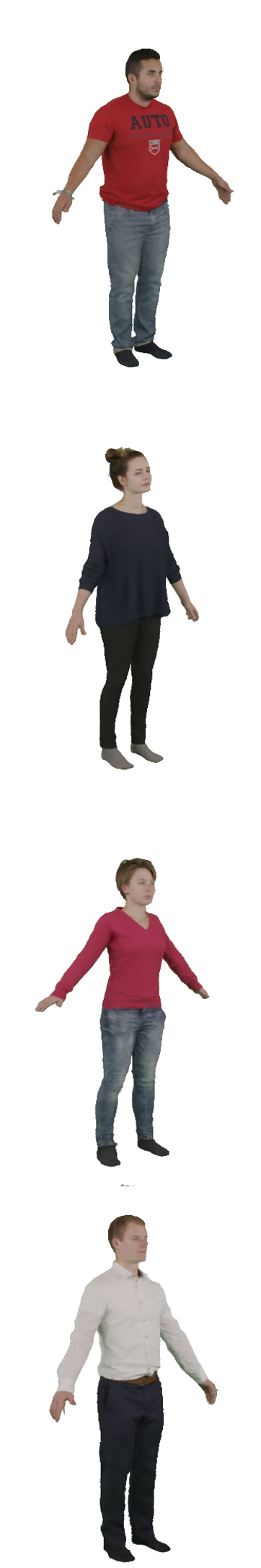}
	\end{minipage}
	}
	\subfigure[view 2]{
	\begin{minipage}[b]{0.10\linewidth}
	\includegraphics[width=1.0\linewidth, trim=0 0 0 0,clip]{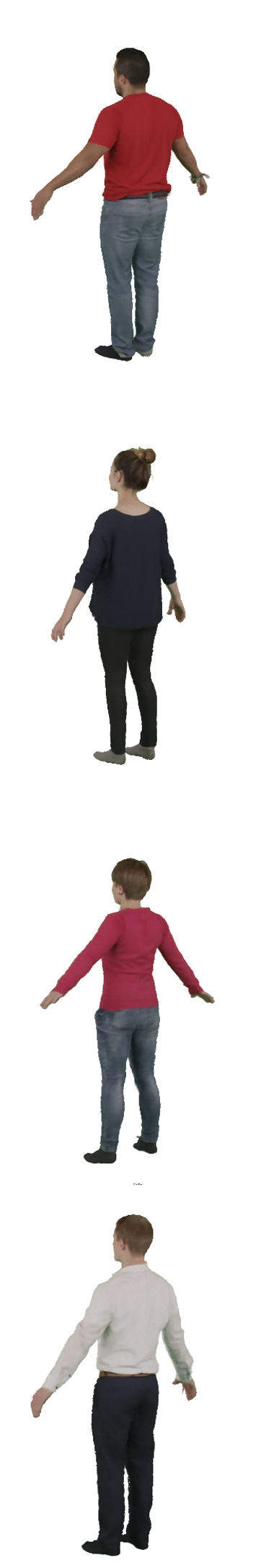}
	\end{minipage}
	}
    \caption{Results of Novel View Synthesis on iPER (a-d) and People-Snapshot (e-h). Our method can synthesize realistic and multi-view consistent results from different camera views while maintaining the subject pose fixed.}
    \label{fig:Novel_View}
\end{figure*}


\setlength{\tabcolsep}{2.8pt}
\begin{table*}[ht]
    \caption{Quantitative comparison of novel view synthesis on People-Snapshot\cite{Video_avatars} and iPER\cite{LWGAN}.}
	\label{Albation}
	\begin{center}
		\renewcommand{\arraystretch}{1.1}
		\begin{tabular}{|c|ccccc|ccccc|ccccc|}
			\hline
			\multirow{2}{*}{Subject ID} & \multicolumn{5}{c|}  {PSNR$\uparrow$} & \multicolumn{5}{c|}{SSIM$\uparrow$} &
			\multicolumn{5}{c|}{LIPIS$\downarrow$} \\
			\cline{2-16}
			& NeRF & SMPLpix & NB & NeRF+U & OURS & NeRF &SMPLpix & NB & NeRF+U & OURS & NeRF &SMPLpix & NB & NeRF+U & OURS \\
			\hline
			male-3-casual & 20.64 & 23.74 & 24.94 & 23.88 & \textbf{29.37}  & .8993 & .9229 & .9428 & .9329 & \textbf{.9703} & .1008 & .0222 & .0326 & .0438 & \textbf{.0168} \\
			male-4-casual & 20.29 & 22.43 & 24.71 & 23.13 & \textbf{28.37}  & .8803 & .9095 & .9469 & .9276 & \textbf{.9605} & .1445 & .0305 & .0423 & .0554 & \textbf{.0268} \\
			female-3-casual & 17.43 & 22.33 & 23.87 & 22.45 & \textbf{28.91} & .8605 & .9288 & .9504  & .9413  & \textbf{.9743} & .1696 & .0270 & .0346 & .0498 & \textbf{.0215} \\
			female-4-casual & 17.63 & 23.35 & 24.37 & 23.13 & \textbf{28.90} & .8578 & .9258 & .9451 & .9276 & \textbf{.9678} & .1827 & .0239 & .0382 & .0556 & \textbf{.0174} \\
			\hline
			iper-009-4-1 & 19.54 & 20.25 & 25.46 & 21.56 & \textbf{30.23} & .7870 & .9018 & .9378 & .8667 & \textbf{.9466} & .2641 & \textbf{.0293} & .0558 & .1197 & .0335 \\
			iper-023-1-1 & 17.41 & 19.48 & 25.44 & 20.25 & \textbf{27.26} & .7623 & .8945 & .9330 & .8656 & \textbf{.9457} & .2769 & .0442 & .0493 & .1109 & \textbf{.0285} \\
			iper-002-1-1 & 16.01 & 19.64 & 23.06 & 18.75 & \textbf{26.99} & .7500 & .8886 & .9394 & .8708 & \textbf{.9502} & .3363 & .0392 &.0476 & .1205 & \textbf{.0285} \\
			iper-026-1-1 & 17.09 & 19.03 & 23.77 & 18.48 & \textbf{26.85} & .7580 & .8574 & .9351 & .8623 & \textbf{.9542} & .2928 & .0494 & .0550 & .1282 & \textbf{.0315} \\
			\hline
		\end{tabular}
	\end{center}
\end{table*}

\subsection{Applications}
The proposed approach learns an animatable NeRF, allowing us to reconstruct the implicit neural representation of the geometry and appearance of the human body, from a monocular video of a person turning around before a camera while holding the A-pose. For the original NeRF~\cite{NeRF}, novel view images (Sec.~\ref{NVS}) can be rendered through volume rendering, and the surface geometry (Sec.~\ref{3DR}) of the scene can be extracted with the Marching Cubes algorithm~\cite{MCubes}.
Since we have explicitly incorporated the SMPL model into the NeRF training process, we can deform the neural radiance field to desired poses for rendering by our pose-guided deformation. This makes our NeRF \textit{animatable}, and thus a new application that can demonstrate new poses or animating the reconstructed people (Sec.~\ref{NPS}) is enabled as shown in Fig.~\ref{fig:motion_transfer} and Table~\ref{novel_pose_synthesis_with_neuralbody}.


\section{Experiments}

\subsection{Implementation Details}
Following NeRF~\cite{NeRF}, we use coarse and fine networks to represent the human body, and use $64$ coarse and $64+32$ fine rays samples for all experiments. 
Focusing on the foreground subject, we set $90\%$ of the rays to be sampled from the foreground, and the remaining $10\%$ to be sampled from the background. 
We set the hyper-parameters as $\left|\mathcal{N}(i)\right|=4$, $\delta=0.2$, $\lambda_1=0.001$, $\lambda_2=0.01$ and $\lambda_d=0.1$. 
We use $512 \times 512$ image in all experiments. 
For training the model, we adopt the Adam optimizer~\cite{Adam}, and it spends about 13 hours on 2 Nvidia GeForce RTX 3090 24GB GPUs.

\begin{figure*}[ht]
    \centering
    \subfigure[Input]{
	\begin{minipage}[b]{0.12\linewidth}
	\includegraphics[width=1.0\linewidth, trim=0 0 0 0,clip]{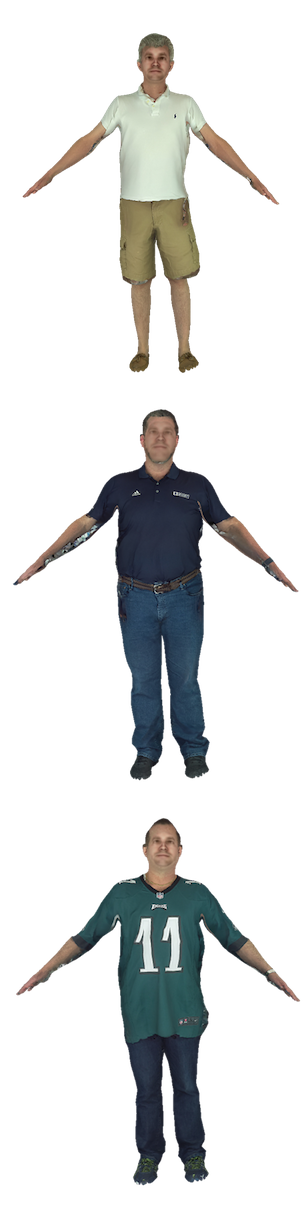}
	\end{minipage}
	}
	\subfigure[NeRF]{
	\begin{minipage}[b]{0.12\linewidth}
	\includegraphics[width=1.0\linewidth, trim=0 0 0 0,clip]{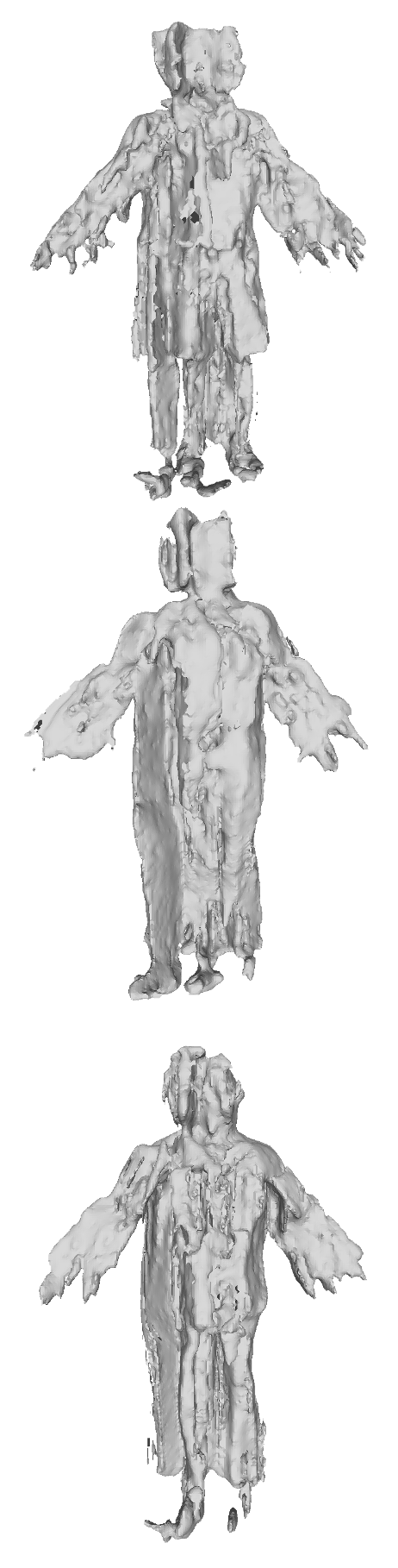}
	\end{minipage}
	}
	\subfigure[NeRF+L]{
	\begin{minipage}[b]{0.12\linewidth}
	\includegraphics[width=1.0\linewidth, trim=0 0 0 0,clip]{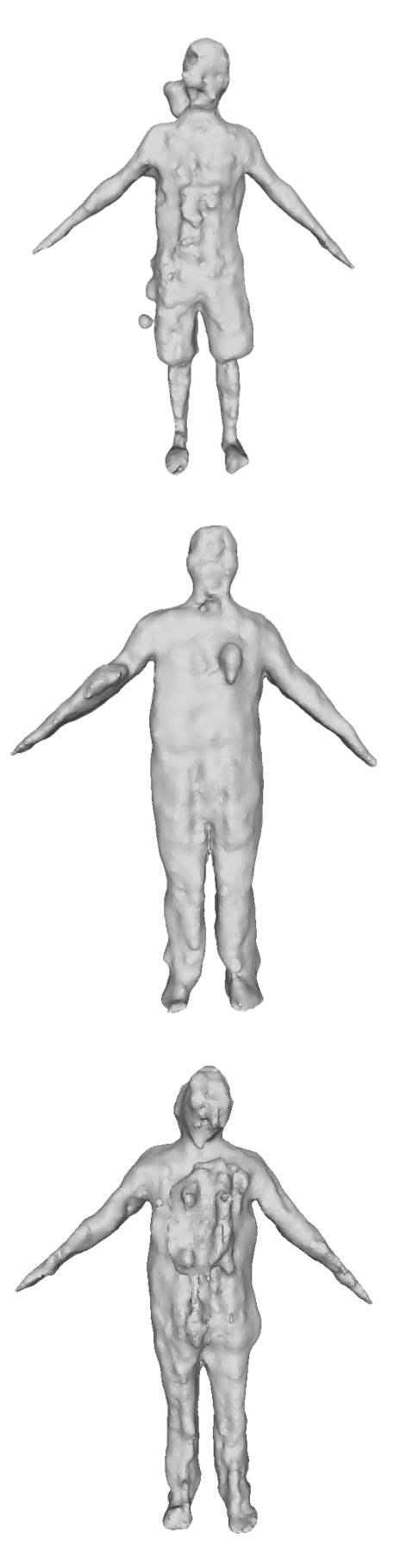}
	\end{minipage}
	}
	\subfigure[NeRF+U]{
	\begin{minipage}[b]{0.12\linewidth}
	\includegraphics[width=1.0\linewidth, trim=0 0 0 0,clip]{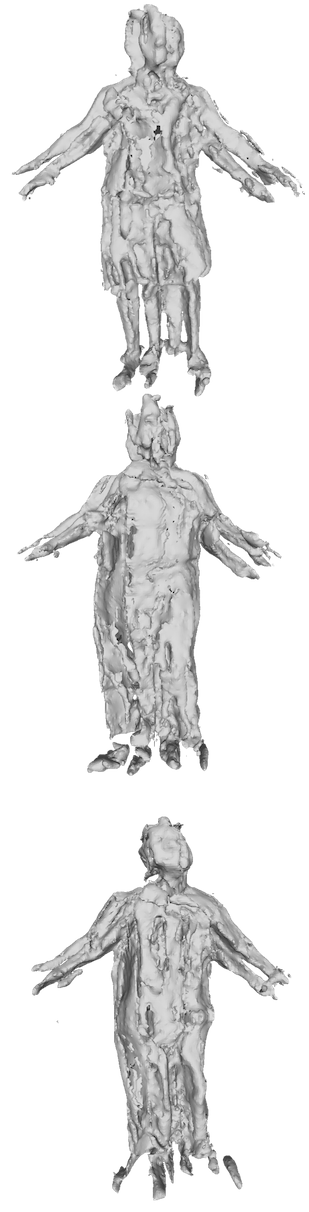}
	\end{minipage}
	}
	\subfigure[Ours]{
	\begin{minipage}[b]{0.12\linewidth}
	\includegraphics[width=1.0\linewidth, trim=0 0 0 0,clip]{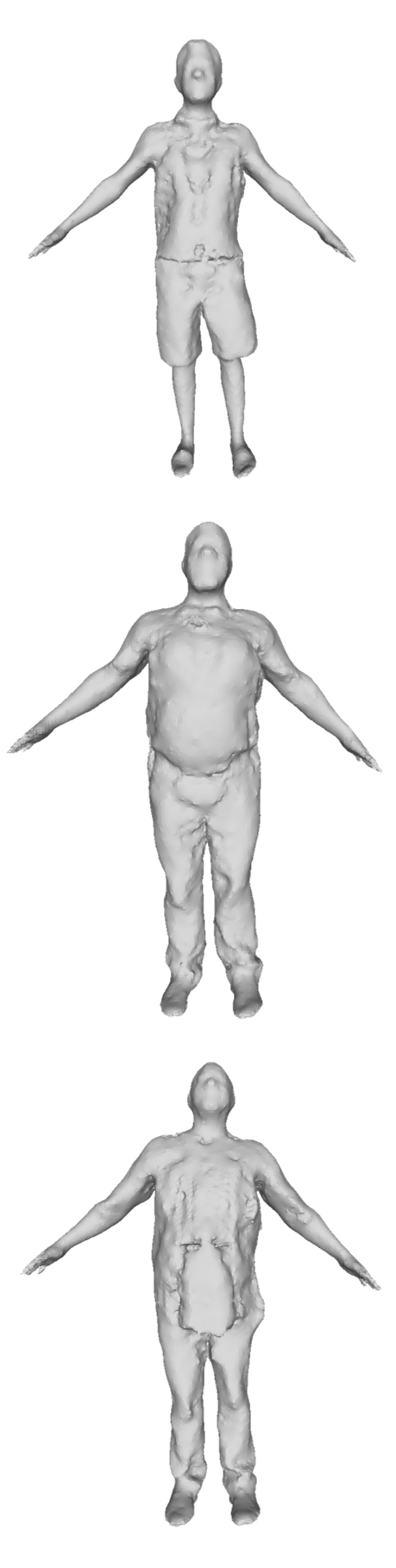}
	\end{minipage}
	}
	\subfigure[NeRF+U(GT)]{
	\begin{minipage}[b]{0.12\linewidth}
	\includegraphics[width=1.0\linewidth, trim=0 0 0 0,clip]{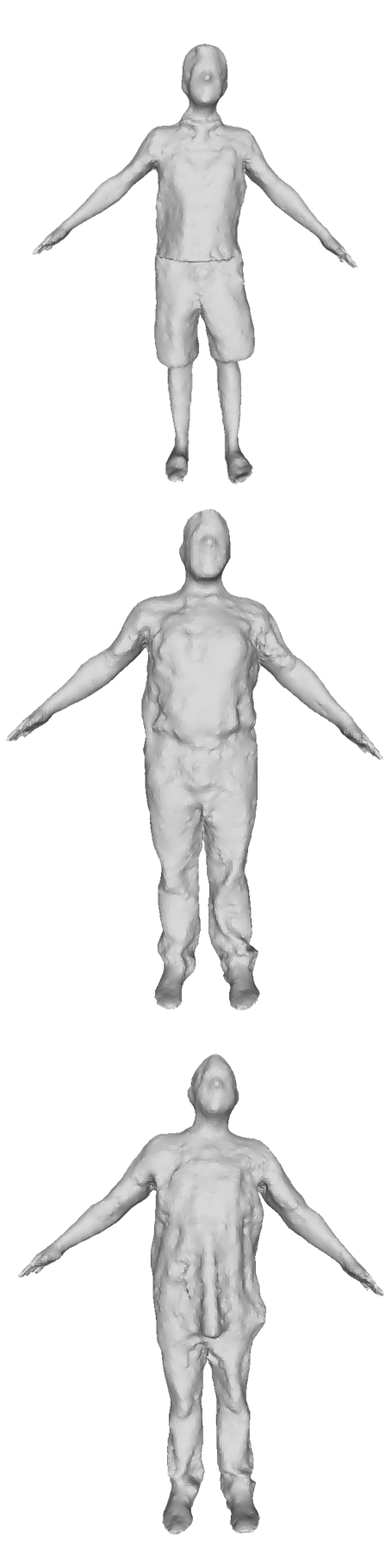}
	\end{minipage}
	}
	\subfigure[GT]{
	\begin{minipage}[b]{0.12\linewidth}
	\includegraphics[width=1.0\linewidth, trim=0 0 0 0,clip]{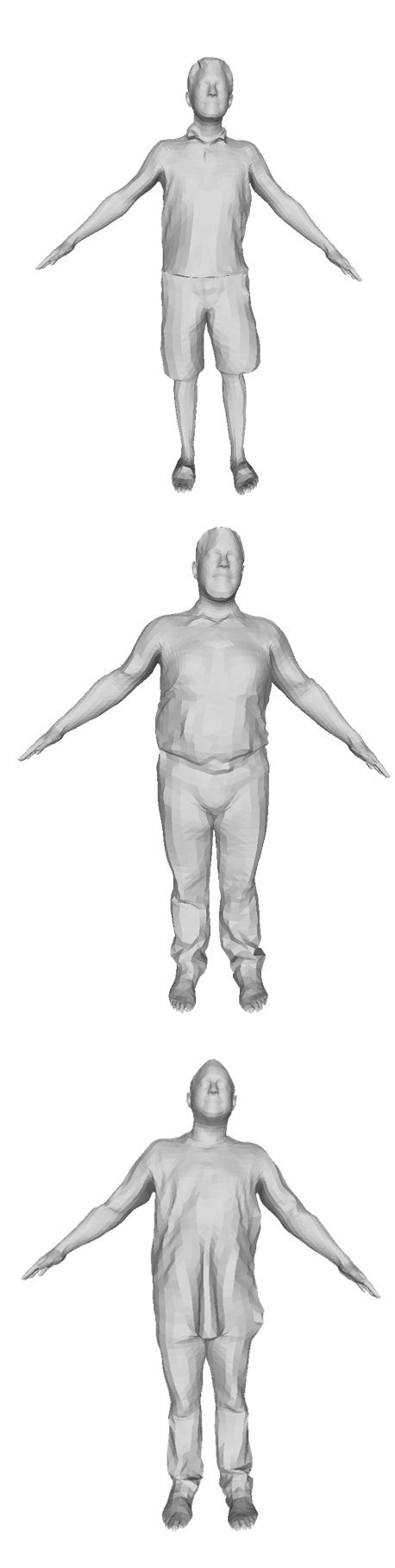}
	\end{minipage}
	}
    \caption{Visualization of 3D reconstruction on Multi-Garment. NeRF\cite{NeRF} and NeRF+U (NeRF + Unpose) fail to reconstruct 3D geometry due to the movement of the subject and the inaccurate SMPL. Compared with NeRF+L (NeRF + Latent) which produces over-smooth or under-smooth results, our results are more reasonable. As a reference, NeRF+U(GT) uses GT SMPL and learns geometry with very high precision, demonstrating the effectiveness of our pose-guided deformation and showing the importance of obtaining accurate SMPL for 3D reconstruction tasks.}
    \label{fig:3D_Rec_on_mutil_garment}
\end{figure*}

\setlength{\tabcolsep}{2.8pt}
\begin{table*}[ht]
    \caption{Quantitative comparison of 3D reconstruction on Multi-Garment.}
	\label{3D_Rec_on_mutil-garment}
	\begin{center}
		\renewcommand{\arraystretch}{1.1}
		\begin{tabular}{|c|ccccc|ccccc|}
			\hline
			\multirow{2}{*}{Subject ID} & \multicolumn{5}{c|}  {P2S$\downarrow$} & \multicolumn{5}{c|}{Chamfer$\downarrow$} \\
			\cline{2-11}
			&NeRF &NeRF+L &NeRF+U &OURS &NeRF+U(GT) &NeRF &NeRF+L &NeRF+U &OURS &NeRF+U(GT) \\
			\hline
			people1 & 65.53 & 13.57 & 33.51 & \textbf{4.09} & 0.86  & 89.32 & 13.96 & 41.81 & \textbf{4.25} & 0.25  \\
			people2 & 36.26 & 11.67 & 28.50 & \textbf{1.55} & 0.85 & 34.95 & 10.78 & 28.86 & \textbf{0.96} & 0.25 \\
			people3 & 34.78 & 16.01 & 36.40 & \textbf{4.17} & 1.17 & 33.62 & 13.83 & 38.36 & \textbf{3.30} & 0.43 \\
			people4 & 33.29 & 26.84 & 32.74 & \textbf{3.53} & 1.06 & 33.70 & 26.59 & 32.08 & \textbf{2.68} & 0.36 \\
			\hline
			Average & 42.46 & 17.02 & 33.28 & \textbf{3.32} & 0.99 & 47.90 & 16.29 & 34.79 & \textbf{2.80}  & 0.32 \\
			\hline
		\end{tabular}
	\end{center}
\end{table*}

\subsection{Datasets and Evaluation}

\noindent \textbf{Datasets}.
To evaluate the effectiveness of the proposed method, we conduct experiments on 3 different datasets, including People-Snapshot~\cite{Video_avatars}, iPER~\cite{LWGAN}, and Multi-Garment~\cite{MGNet}.
People-Snapshot\cite{Video_avatars} and iPER\cite{LWGAN} datasets both contain different monocular RGB videos captured in real-world scenes, where the subjects hold an A-pose and turn around before a fixed camera. 
In addition, iPER dataset also contains videos of the same person with random motion sequences. Multi-Garment~\cite{MGNet} dataset contains 3D scanned human body models and textures and the corresponding registered SMPLD models that can be used for animation.
We selected 4 human body models to synthesize the videos, according to motion sequences which the subjects rotate while holding an A-pose in People-Snapshot dataset.
People-Snapshot and iPER datasets are mainly used for the evaluation of novel view synthesis and novel pose synthesis experiments.
And the synthetic data from Multi-Garment dataset are used to evaluate the quality of the 3D reconstructions.

\noindent \textbf{Evaluation}. 
In our experiments, we use A-pose frames (2 circles) for training, and the remaining A-pose frames (1 circle) for testing novel view synthesis and random pose frames for testing novel pose synthesis. 
Since there are depth and scale ambiguities in optimizing the SMPL parameters for monocular videos, 
we also optimize the SMPL parameters of the test frames for quantitative evaluation. 
Note that the parameters of the neural radiance field network remain fixed. 
For quantitative evaluation, we evaluate our method for novel view synthesis and novel pose synthesis using the following metrics: peak signal-to-noise ratio (PSNR), structural similarity index (SSIM~\cite{SSIM}), and learned perceptual image patch similarity (LPIPS~\cite{lpips}). 
For 3D reconstruction, we use point-to-surface Euclidean distance (P2S) and Chamfer distance~\cite{Chamfer} (in cm) between the reconstructed and the ground truth surfaces. 
We register our meshes to ground truth geometry for comparison in consideration of scale and depth ambiguities. 
The datasets from the real scenes don't have the corresponding ground truth geometry, and we only provide qualitative results.

\begin{figure}[ht]
    \centering
    \subfigure[Input]{
	\includegraphics[height=0.72\linewidth, trim=0 0 0 0,clip]{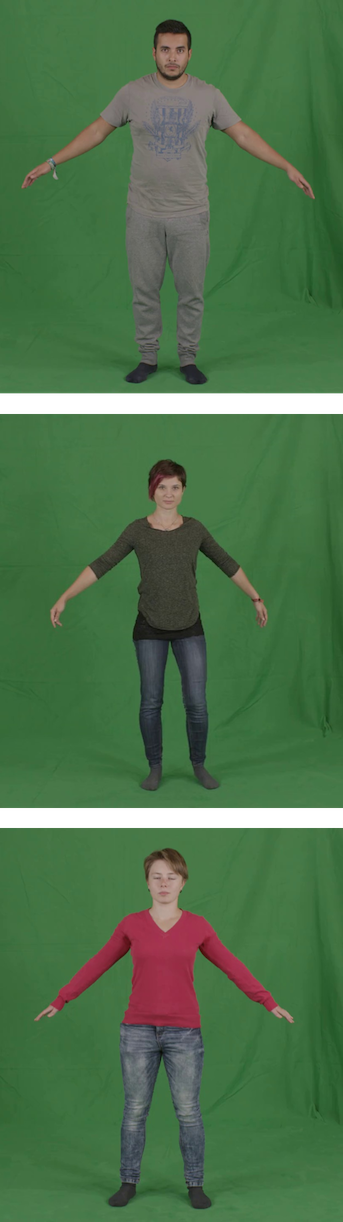}
	}
	\subfigure[Ours]{
	\includegraphics[height=0.72\linewidth, trim=10 0 20 0,clip]{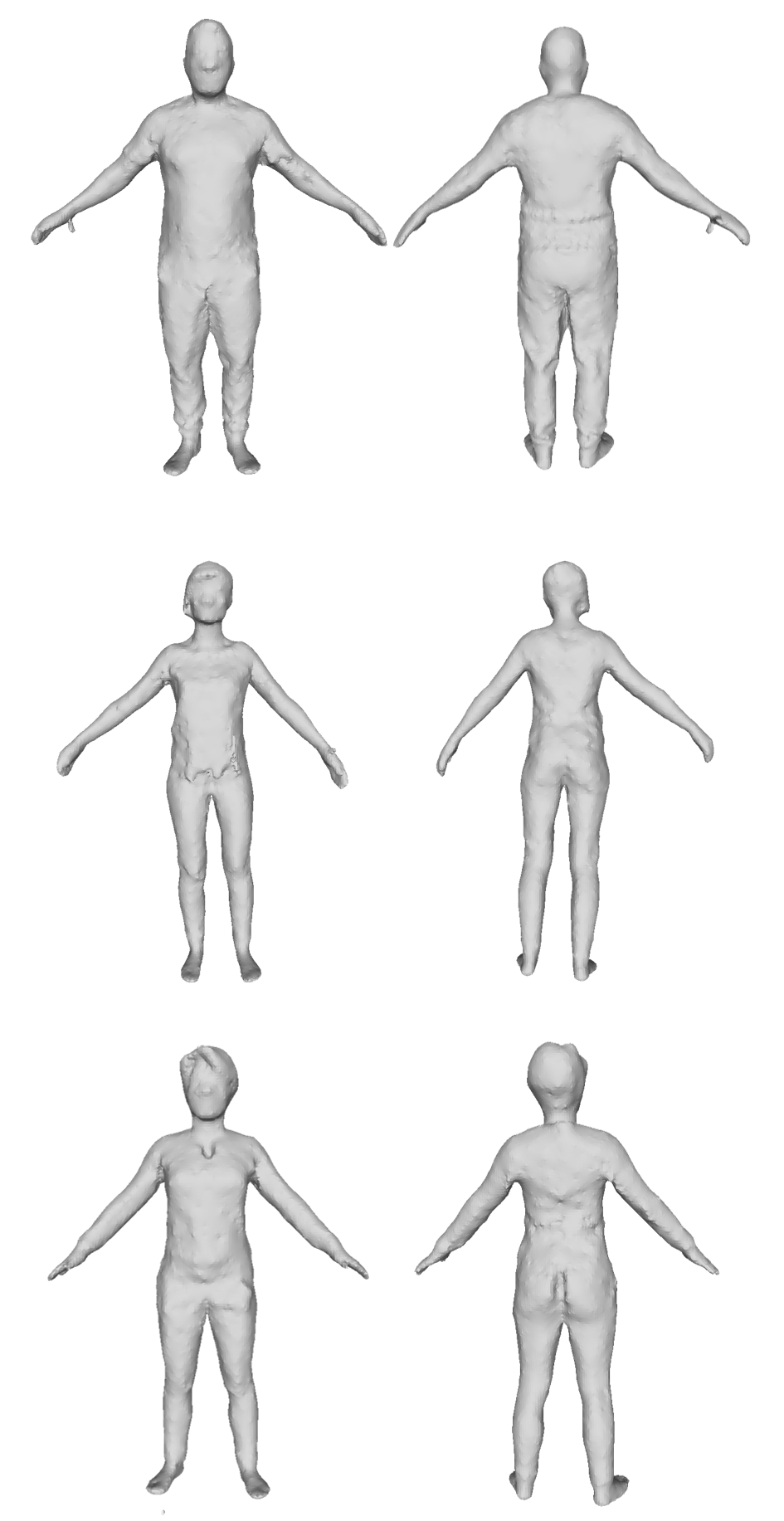}
	}
	\subfigure[Video Avatars]{
	\includegraphics[height=0.72\linewidth, trim=20 0 0 0,clip]{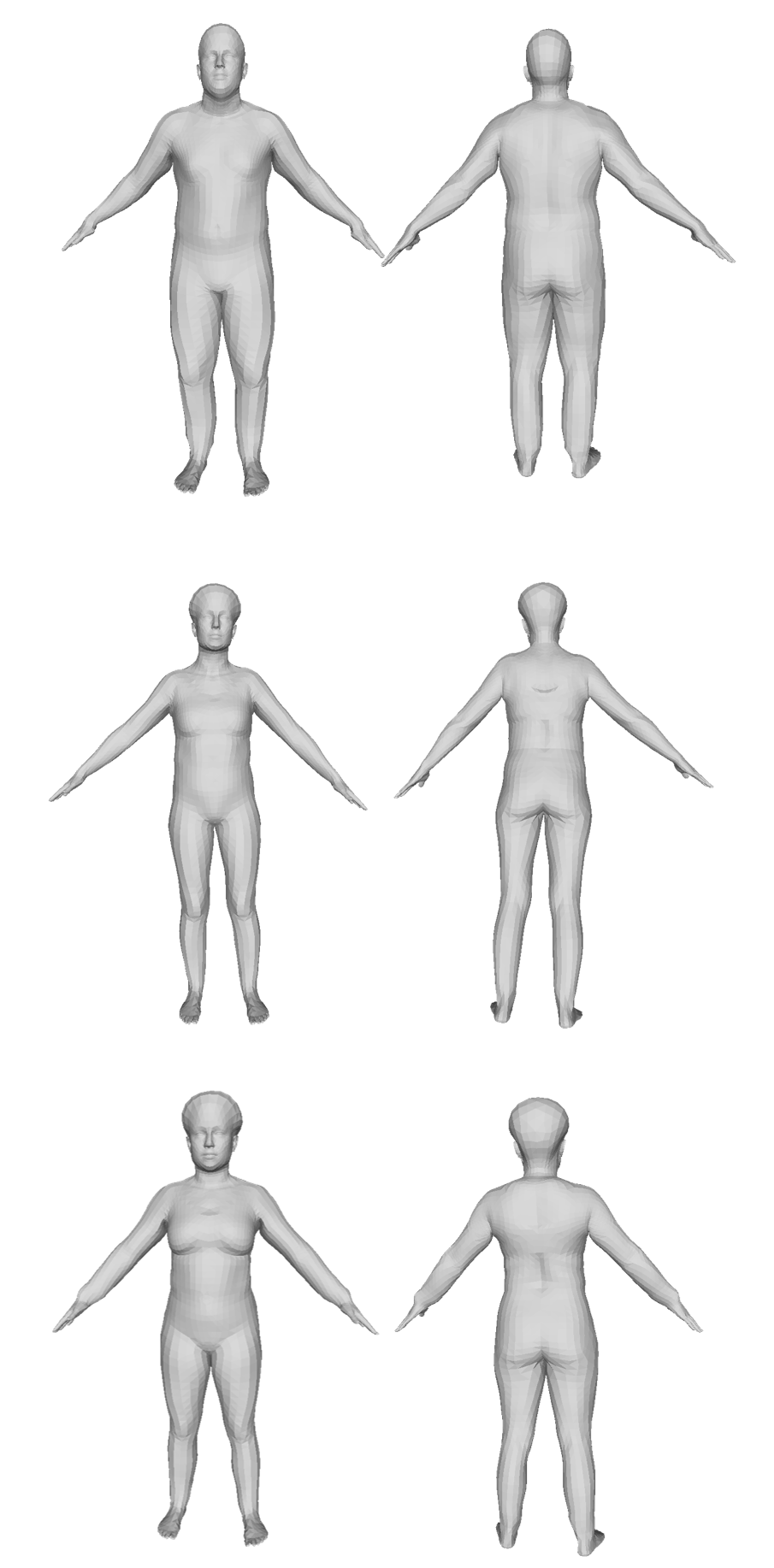}
	}
    \caption{Comparisons of 3D reconstruction results on People-Snapshot with video avatars~\cite{Video_avatars}. Compared with Video Avatars\cite{Video_avatars}, our approach can generate more details such as hairs and clothes wrinkles.}
    \label{fig:3D_Rec_with_SMPLD}
\end{figure}

\begin{figure}[ht]
    \centering
    \includegraphics[width=1.0\linewidth]{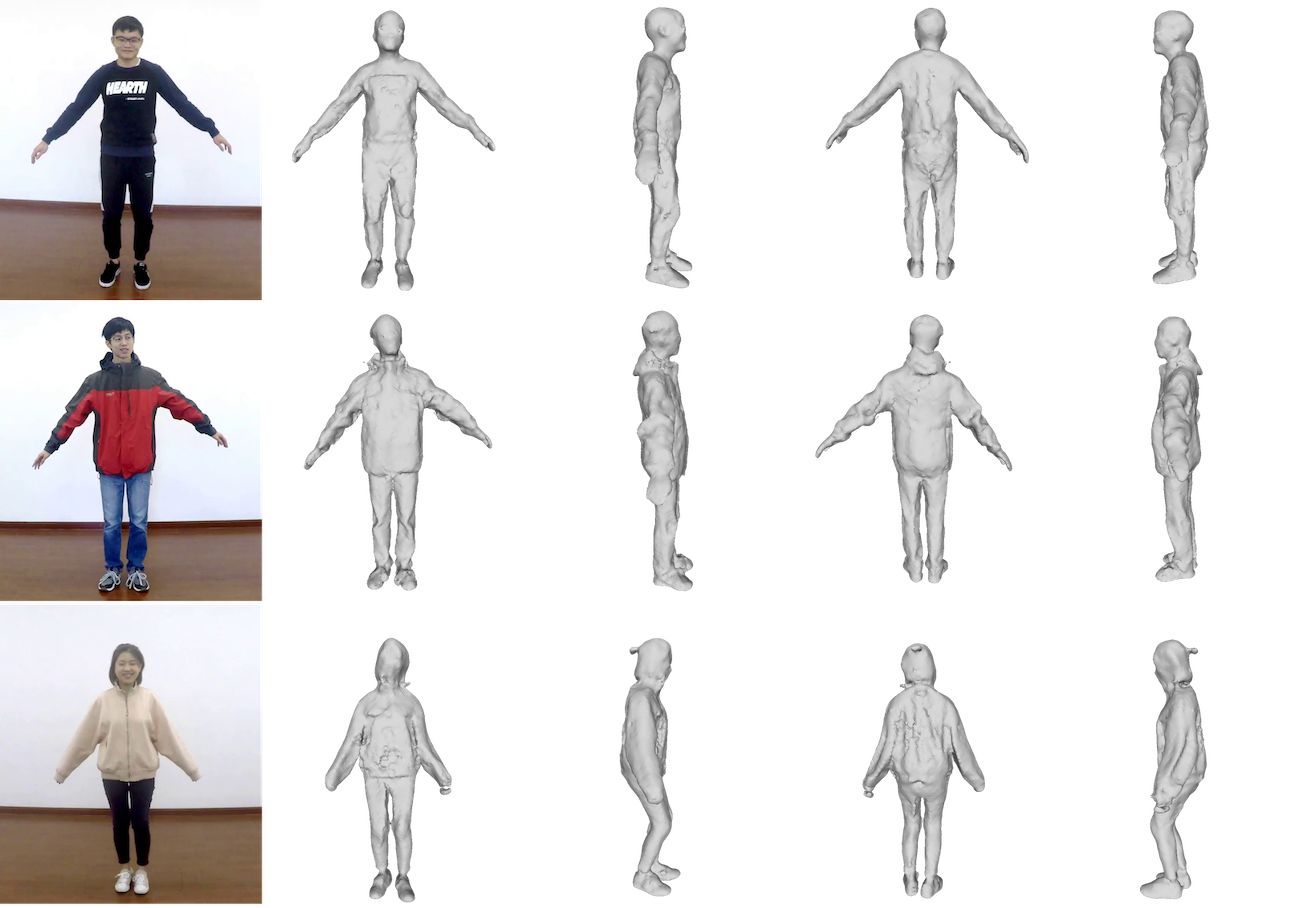}
    \caption{Visualization of our reconstructed geometry on iPER from different views.}
    \label{fig:3D_Rec_iPER}
\end{figure}

\begin{figure}[ht]
    \centering
    \includegraphics[width=1.0\linewidth]{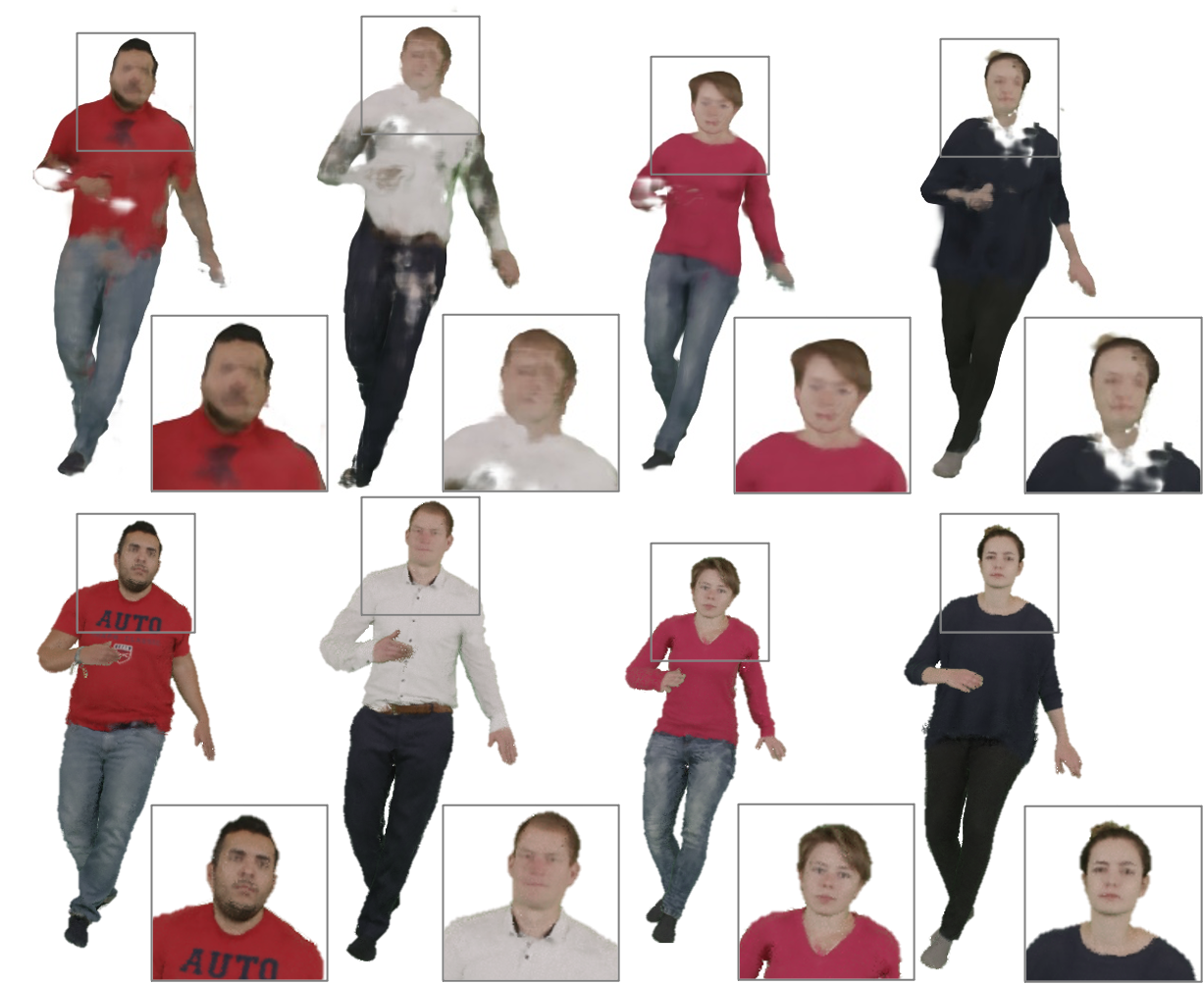}
    \caption{Comparisons between NeuralBody\cite{Neural_Body} (first row) and Ours (second row) on novel pose synthesis task. In contrast to NeuralBody, which fails to synthesize novel poses, our approach generalizes better on novel poses that are very different from the training poses.}
    \label{compare_with_neuralbody_on_unseen_pose_in_detail}
\end{figure}

\subsection{Novel view synthesis}
\label{NVS}
Like the original NeRF~\cite{NeRF}, our animatable NeRF can be rendered from arbitrary views (of the same pose). 
Since the monocular video does not have corresponding novel view images, we use the first part of the A-pose video frames to train our model and the remaining frames to test the rendered novel view images\footnote{Technically, this is not a ``novel'' view apart from slightly different human pose}.
To compare against the original NeRF, we first transform the global rotation and translations of the SMPL estimation from every video frame to camera pose as if we are handling a collection of multi-view images of an almost static scene since the subject is holding an A-pose.
In order to verify the effectiveness of our pose fine-tuning strategy, we first use VIBE~\cite{VIBE} to estimate the parameters of SMPL, and conduct comparative experiments between ``Ours'' (NeRF + Unpose + Pose Refinement) and ``NeRF+U'' (NeRF + Unpose) as shown in Fig.~\ref{fig:Comparision_with_different_method}.
We also compare the proposed method with several state-of-the-art (SOTA) methods, including NeuralBody\cite{Neural_Body}(NB) and SMPLpix\cite{smplpix}.
NeuralBody, which also combines SMPL and NeRF, is able to reconstruct dynamic human bodies from monocular video.
SMPLpix takes SMPL pose as input to generate images via a neural rendering network.
Table~\ref{Albation} quantitatively compares the results of different approaches about novel view synthesis on the People-Snapshot and iPER datasets. 
As described in the table, our proposed approach achieves higher PSNR and SSIM scores compared to other approaches. 
We also provide qualitative comparisons in Fig.~\ref{fig:Comparision_with_different_method} with the person examples drawn from the People-Snapshot the iPER datasets. 
We can see that the proposed approach produces more realistic and reliable results.
NeRF fails to handle such dynamic scenes since the movement of the subject violates the multi-view consistency requirement. 
Experiments in Fig.~\ref{fig:Comparision_with_different_method}(d) and Table~\ref{Albation} (see NeRF+U) also show that inaccurate SMPL parameters cause a very negative impact. 
In contrast, after taking pose refinement into consideration, the quality of novel view synthesis has been significantly improved. 
As shown in Fig.~\ref{fig:Novel_View}(b)(c), SMPLpix and NeuralBody seem to overfit the training frames, while our results better preserve the details such as faces and hands.
Fig.~\ref{fig:Novel_View} shows the realistic rendering results of more views of the proposed method on more persons with different dresses and hairstyles, indicating the applicability and robustness of the proposed method in real scenarios.


\subsection{3D human reconstruction}
\label{3DR}
On this task, we compare against the original NeRF~\cite{NeRF} and NeRF+L baselines. NeRF+L extends NeRF to condition it on a (per-frame) learnable latent deformation code to handle dynamic scenes as shown in Fig~\ref{fig:3D_Rec_on_mutil_garment}(c) and Table~\ref{3D_Rec_on_mutil-garment}.
For synthetic data, we also show results using ground truth SMPL parameters (NeRF+U(GT)) for pose-guided deformation as the upper bound as shown in Fig~\ref{fig:3D_Rec_on_mutil_garment}(f) and Table~\ref{3D_Rec_on_mutil-garment}.
Quantitative compassion of different strategies in 3D human reconstruction is shown in Table~\ref{3D_Rec_on_mutil-garment}. 
We can see that the proposed approach achieves much lower P2S and Chamfer distances, demonstrating the superiority of the proposed approach in reconstructing accurate 3D geometry. 
Fig.~\ref{fig:3D_Rec_on_mutil_garment} compares the qualitative results of 3D reconstruction. 
We can see that NeRF fails to learn reasonable 3D geometry of the human subject with \textit{small} movements. 
NeRF+U(see Sec. \ref{NVS}) also produces messy results. Compared with NeRF+L, which produces over-smooth or under-smooth results, the proposed approaches better capture the geometric details such as cloth wrinkles, faces, and hairs. 
With ground truth SMPL parameters, the P2S and Chamfer distance are much lower than all the approaches, which demonstrates the necessity of obtaining accurate poses and the effectiveness of our approximated pose-guided deformation. 
In Fig.~\ref{fig:3D_Rec_with_SMPLD}, we compare the reconstruction results with video avatar~\cite{Video_avatars}, which deforms vertices of the SMPL model to fit the 2D human silhouettes over the video sequence. 
We can see that the implicit learning of subject geometry with animatable NeRF generates a better quality of details, including cloth wrinkles, hair, and accessories. 
In Fig.~\ref{fig:3D_Rec_iPER} we show the reconstruction results of persons with varied clothes and hairstyles from iPER dataset. 
Although our pose-guided deformation does not take the deformation of clothes into account, the proposed method is capable of capturing the high-quality 3D geometry details, such as the hood (second line) and pigtail (third line), as long as the clothes do not have violent deformation.


\begin{figure*}[ht]
    \centering
    \includegraphics[width=1.0\linewidth]{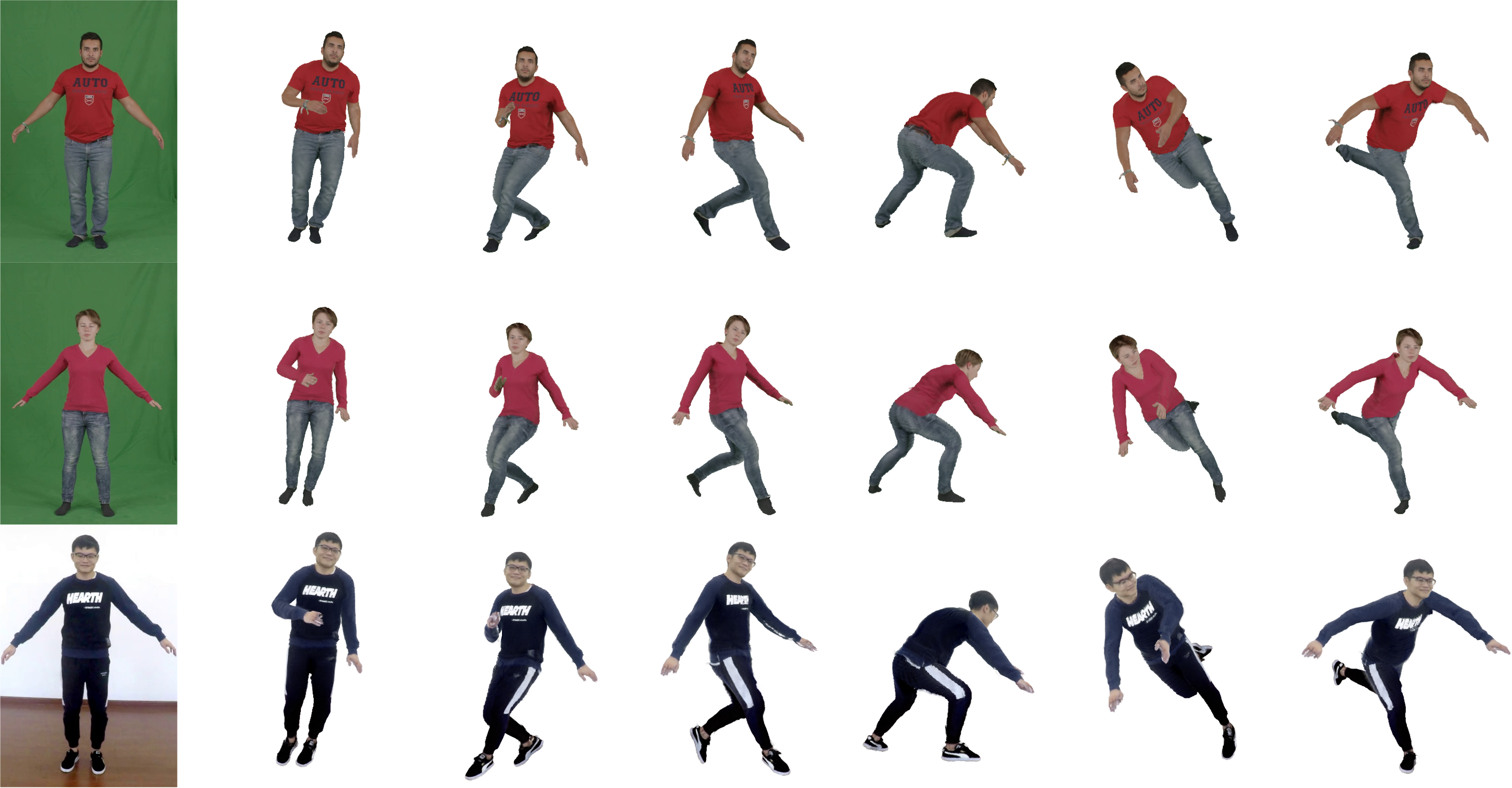}
    \caption{Novel pose synthesis on People-Snapshot\cite{Video_avatars} and iPER\cite{LWGAN}. We can feed novel SMPL pose parameters to the trained animatable NeRF to synthesize novel pose images. 
    Although trained only on A-pose images, our animatable NeRF has the capability to stably render new images containing complex poses.
    }
    \label{fig:motion_transfer}
\end{figure*}

\subsection{Novel pose synthesis} 
\label{NPS}
Due to our explicit control of deformation via SMPL, our method can synthesize images under unseen poses even with only simple A-pose sequences as input. 
To quantitatively evaluate the capability of our method on novel pose synthesis, we train the model using A-pose videos and test it using random pose videos of the same person. 
NeuralBody\cite{Neural_Body} is the most similar work to ours in the sense that it also combines NeRF with SMPL. 
Compared to ours, it handles complex cloth geometry (which is not modeled by SMPL) better due to its use of latent code. 
However, each vertex's latent code would affect a much larger region after several sparse convolution layers, resulting in unpredictable artifacts for novel pose synthesis (see Fig. \ref{compare_with_neuralbody_on_unseen_pose_in_detail}).
Table~\ref{novel_pose_synthesis_with_neuralbody} shows that our method achieves much better results than NeuralBody on novel pose synthesis. 
Qualitative visualizations of novel pose synthesis on People-snapshot and iPER are provided in Fig.~\ref{fig:motion_transfer}. 
Specifically, different poses are fed into the trained animatable NeRF to obtain the aforementioned renderings. 
Despite the significant differences between the test novel poses and the training poses, the results show that our method can still produce realistic images with well preserved identity and cloth details of the subjects.

\setlength{\tabcolsep}{2.8pt}
\begin{table}[ht]
    \caption{Quantitative comparison about novel pose synthesis with NeuralBody(NB)\cite{Neural_Body} on the iPER dataset.}
	\label{novel_pose_synthesis_with_neuralbody}
	\begin{center}
		\renewcommand{\arraystretch}{1.1}
		\begin{tabular}{|c|cc|cc|cc|}
			\hline
			\multirow{2}{*}{Subject ID} & \multicolumn{2}{c|}  {PSNR$\uparrow$} & \multicolumn{2}{c|}  {SSIM$\uparrow$} &
			\multicolumn{2}{c|}  {LPIPS$\downarrow$} \\
			\cline{2-7}
			& NB & OURS & NB & OURS & NB & OURS \\
			\hline
			iper-009-4-2 & 20.95 & \textbf{24.11} & \textbf{.9035} & .8927 & .0980 & \textbf{.0782} \\
			\hline
			iper-023-1-2 & 20.28 & \textbf{21.98}  & \textbf{.9009} & .8940 & .0870 & \textbf{.0644} \\
			\hline
			iper-026-1-2 & 17.42 & \textbf{19.27} & \textbf{.8795} & .8713 & .1192 & \textbf{.0990} \\
			\hline
			iper-002-1-2 & 19.07 & \textbf{23.47} & .8957 & \textbf{.9165} & .0749 & \textbf{.0483} \\
			\hline
		\end{tabular}
	\end{center}
\end{table}


\section{Discussion}

In the following, we will discuss the proposed approach in details from the following aspects:  analysis of pose refinement (Sec.~\ref{analysis_of_pose_refinement}) and canonical poses (Sec.~\ref{analysis_of_canonical_poses}), Impact of background regularization (Sec.~\ref{impact_of_background_regularization}) and view direction (Sec.~\ref{impact_of_view_direction}).


\subsection{Analysis of Pose Refinement}
\label{analysis_of_pose_refinement}

\begin{figure}[ht]
    \centering
	\subfigure[VIBE est.]{
	\begin{minipage}[b]{0.46\linewidth}
	\includegraphics[width=1.0\linewidth, trim=0 0 0 0,clip]{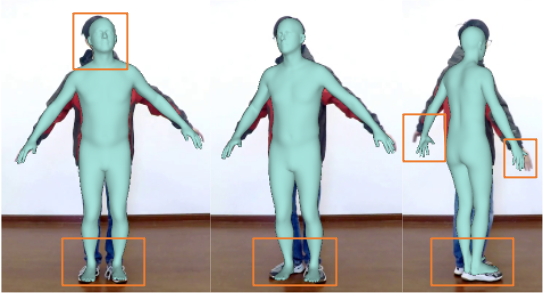}
	\end{minipage}
	}   
	\subfigure[Ours]{
	\begin{minipage}[b]{0.46\linewidth}
	\includegraphics[width=1.0\linewidth, trim=0 0 0 0,clip]{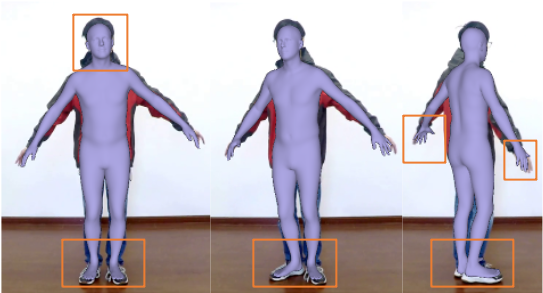}
	\end{minipage}
	} 
    \caption{Visual comparison before and after pose refinement on iPER\cite{LWGAN}. After pose refinement, the SMPL model is better aligned with the input image (e.g. foot joints).
    }
	\label{Pose_Refine_Vis}
\end{figure}

Here we discuss the impact of pose refinement on our approach. 
Our method relies on SMPL parameters for explicit deformation, so inaccurate SMPL estimation may lead to catastrophic results as shown in Fig.~\ref{fig:Comparision_with_different_method}(d).
We initialize the SMPL parameters with estimations from VIBE\cite{VIBE}, which is the state-of-the-art pose and shape estimation method. 
However, pose estimation from monocular videos usually suffers from depth ambiguity.
As shown in Fig.~\ref{Pose_Refine_Vis}(a), although our input video is a simple A-pose, the SMPL estimated by VIBE is usually misaligned at the foot joints.
After pose refinement, the SMPL model is better fitting to the input image as shown in Fig.~\ref{Pose_Refine_Vis}(b).

\subsection{Analysis of Canonical Poses}
\label{analysis_of_canonical_poses}

\begin{figure}[ht]
    \centering
    \subfigure[Input]{
	\begin{minipage}[b]{0.182\linewidth}
	\includegraphics[width=1.0\linewidth, trim=0 0 0 0,clip]{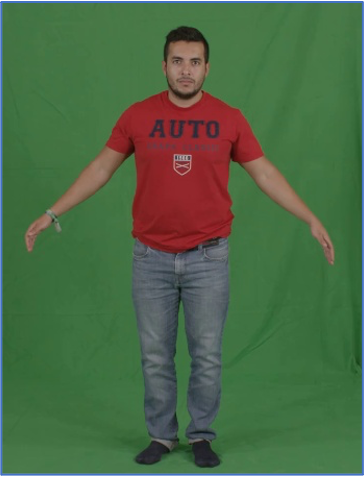}
	\end{minipage}
	}   
	\subfigure[A-pose]{
	\begin{minipage}[b]{0.182\linewidth}
	\includegraphics[width=1.0\linewidth, trim=0 0 0 0,clip]{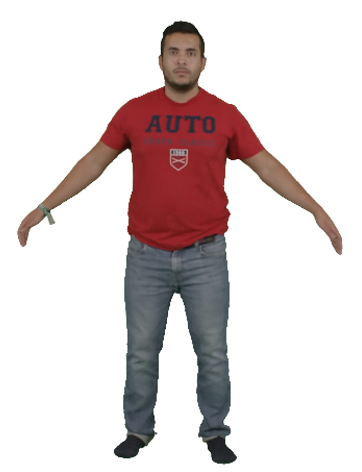}
	\end{minipage}
	}   
	\subfigure[T-pose]{
	\begin{minipage}[b]{0.24\linewidth}
	\includegraphics[width=1.0\linewidth, trim=0 0 0 0,clip]{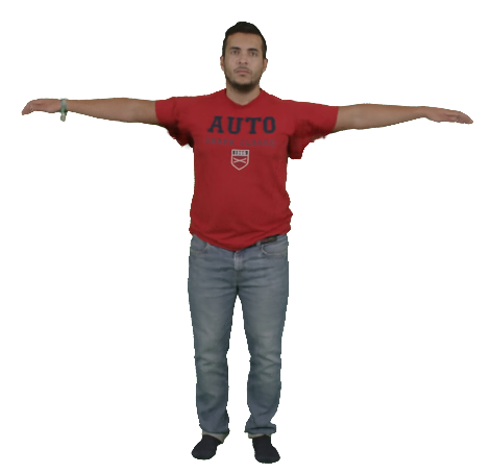}
	\end{minipage}
	} 
    \subfigure[X-pose]{
	\begin{minipage}[b]{0.24\linewidth}
	\includegraphics[width=1.0\linewidth, trim=0 0 0 0,clip]{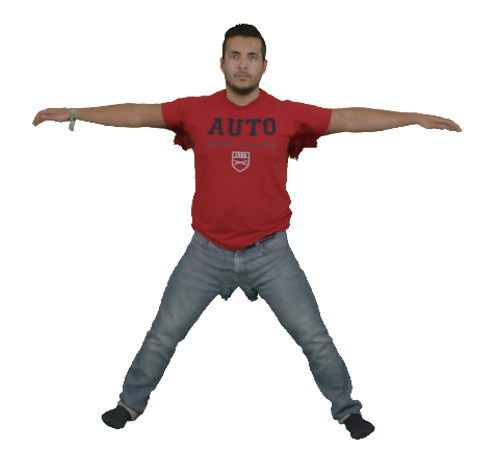}
	\end{minipage}
	} 
    \caption{Visualization of different canonical NeRF spaces with different canonical poses during training on People-Snapshot\cite{Video_avatars}.}
	\label{Canonical_pose}
\end{figure}

In this section, we discuss the effect of using different canonical poses in canonical space on the reconstruction results. 
Since the pose-guided deformation is explicitly based on SMPL\cite{SMPL}, we will get different canonical spaces with different canonical poses. 
Therefore, the choice of canonical poses has a crucial impact on the reconstruction and novel pose synthesis. Here we will discuss the reconstruction results of three different canonical poses: A-pose, T-pose, and X-pose. 
A-pose is the average pose of the body poses of the training frames, which is the closest to the poses in the training frames.  
T-pose is the SMPL model's rest pose, where the arms are far away from the body, but the legs are closer to each other. 
In comparison, our customized X-pose offers more spread body parts (see Fig. \ref{Canonical_pose}(d)).

As shown in Fig. \ref{Canonical_pose}, using A-pose as the canonical pose offers the best quality for canonical space NeRF learning, while using T-pose and X-pose as canonical poses result in some artifacts under the axilla and the thighs. 
If a point is close to two different SMPL body parts (e.g. body and arm, two legs), it is hard to decide which part the point belongs to since SMPL models unclothed human body only without taking the offset of the clothes into consideration.

\begin{figure}[ht]
    \centering
    \subfigure[Input]{
	\begin{minipage}[b]{0.18\linewidth}
	\includegraphics[width=1.0\linewidth, trim=0 0 0 0,clip]{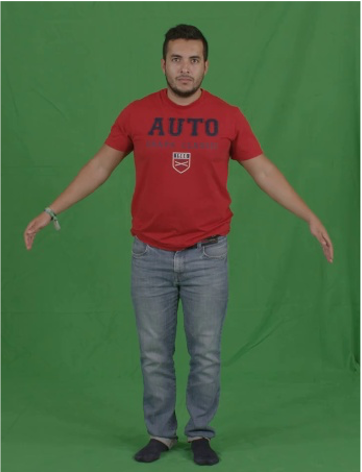}
	\end{minipage}
	}
	\subfigure[A-pose]{
	\begin{minipage}[b]{0.21\linewidth}
	\includegraphics[width=1.0\linewidth, trim=0 0 0 0,clip]{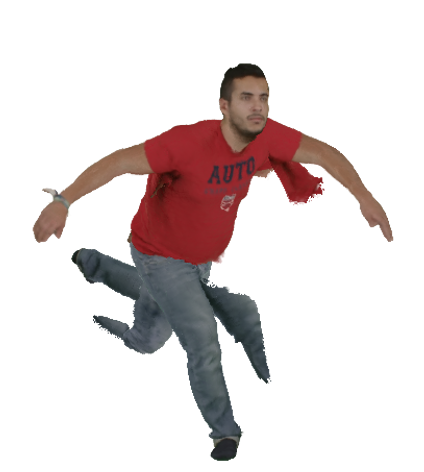}
	\end{minipage}
	}   
	\subfigure[T-pose]{
	\begin{minipage}[b]{0.21\linewidth}
	\includegraphics[width=1.0\linewidth, trim=0 0 0 0,clip]{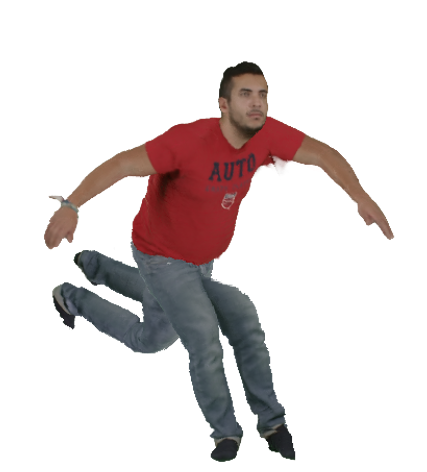}
	\end{minipage}
	} 
    \subfigure[X-pose]{
	\begin{minipage}[b]{0.21\linewidth}
	\includegraphics[width=1.0\linewidth, trim=0 0 0 0,clip]{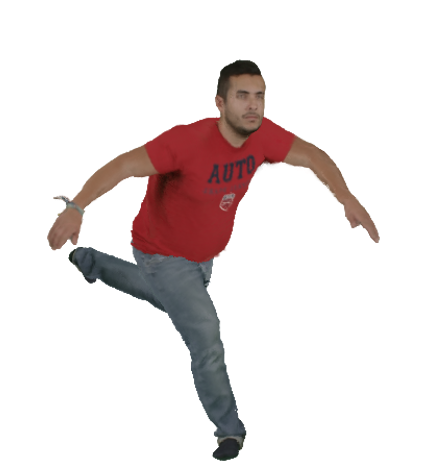}
	\end{minipage}
	} 
    \caption{Novel pose synthesis with different canonical poses on People-Snapshot\cite{Video_avatars}. X-pose is a better choice for novel pose synthesis compared to A-pose and T-pose, which produce unacceptable artifacts (e.g. multiple legs).}
	\label{novel_pose_in_difference_canonical_pose}
\end{figure}

For reconstruction, A-pose is the best choice, but X-pose is more suitable for new pose synthesis. As shown in Fig. ~\ref{novel_pose_in_difference_canonical_pose}. 
When using A-pose and T-pose as canonical poses for synthesizing a pose different from the poses in the training frames, there exist some unacceptable artifacts (e.g. multiple legs). This is because the different body parts are too close to each other in the canonical space of A-pose or T-pose, so that one body part (e.g. left leg) will be deformed by the transformation of another body part (e.g. right leg), resulting in multiple legs in Fig. ~\ref{novel_pose_in_difference_canonical_pose}(b)(c). 

\subsection{Impact of Background Regularization}
\label{impact_of_background_regularization}

\begin{figure}[ht]
    \centering
    \subfigure[Input]{
	\begin{minipage}[b]{0.18\linewidth}
	\includegraphics[width=1.0\linewidth, trim=0 0 0 0,clip]{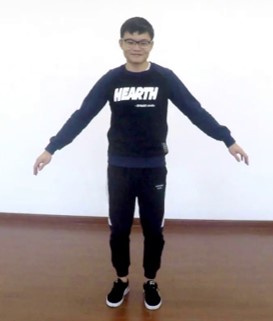}
	\end{minipage}
	}   
	\subfigure[w/o background reg.]{
	\begin{minipage}[b]{0.35\linewidth}
	\includegraphics[width=1.0\linewidth, trim=0 0 0 0,clip]{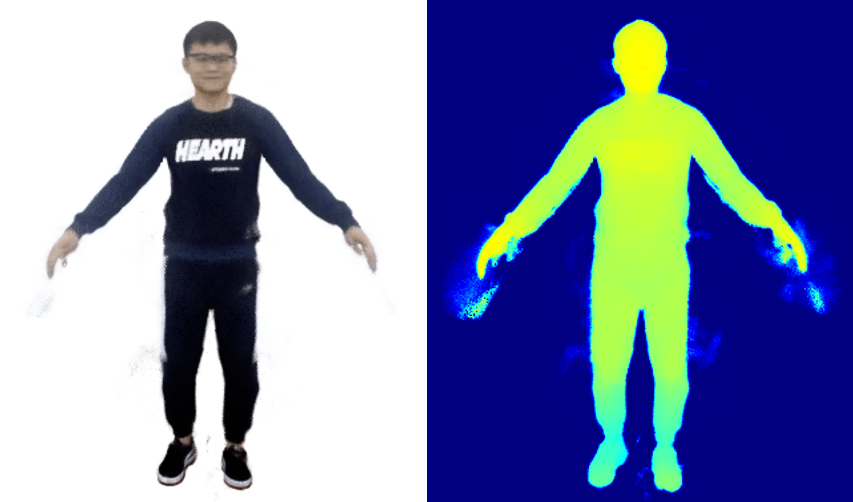}
	\end{minipage}
	} 
	\subfigure[w/ background reg.]{
	\begin{minipage}[b]{0.36\linewidth}
	\includegraphics[width=1.0\linewidth, trim=0 0 0 0,clip]{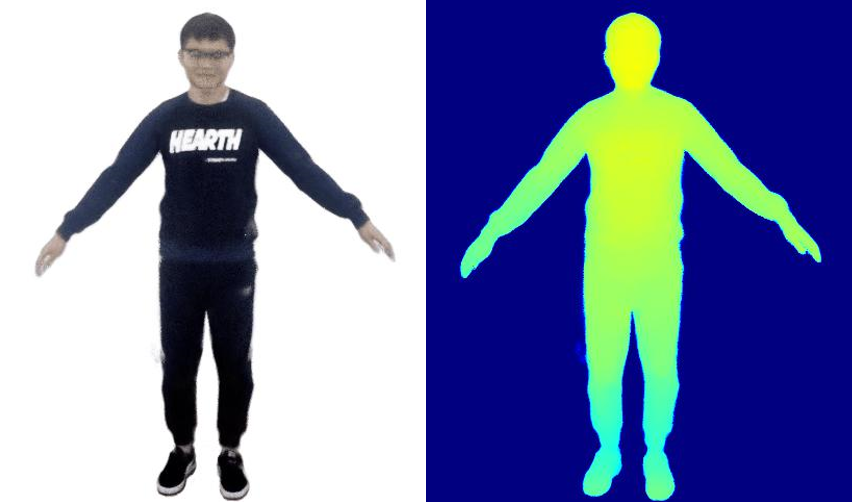}
	\end{minipage}
	}   
    \caption{Impact of background regularization on iPER\cite{LWGAN}. The background regularization can effectively reduce the artifacts in the background region.}
	\label{BG_Regularization}
\end{figure}

In this section, we discuss the benefits of background regularization for our approach. Our approach focuses on the animatable NeRF of the human from a monocular video. 
To avoid possible negative influence from the background, we set the background to white uniformly (with the help of an off-the-shelf segmentation network). 
However, it is common to appear some noisy density regions in the empty space (i.e. non-human space) after training.
As shown in Fig. \ref{BG_Regularization}(b), although the background of the image is white, we notice some noisy non-zero density regions from the depth map. 
This is because there is an ambiguity between the background (white in our case) and the cloth which happens to be the same color as the background. 
To deal with this problem, we introduce background regularization to encourage the density of the background region to be zero. 
With background regularization, the artifacts in the empty space are significantly reduced as shown in Fig.~\ref{BG_Regularization}(c).

\begin{figure}[ht]
    \centering
    \subfigure[Input]{
	\begin{minipage}[b]{0.22\linewidth}
	\includegraphics[width=1.0\linewidth, trim=0 0 0 0,clip]{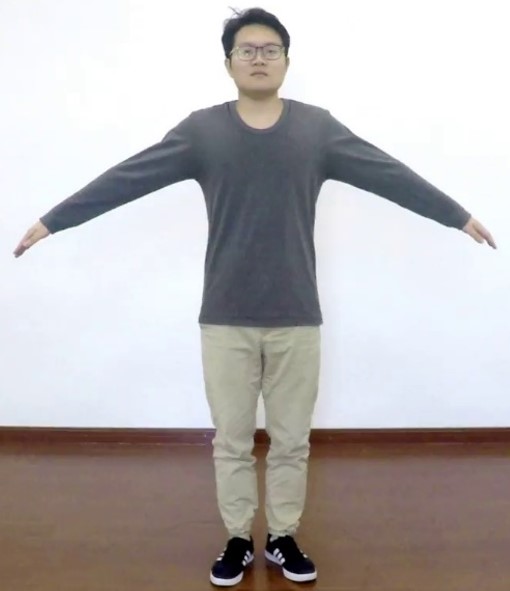}
	\end{minipage}
	}   
	\subfigure[w/ viewing direction]{
	\begin{minipage}[b]{0.33\linewidth}
	\includegraphics[width=1.0\linewidth, trim=0 0 0 0,clip]{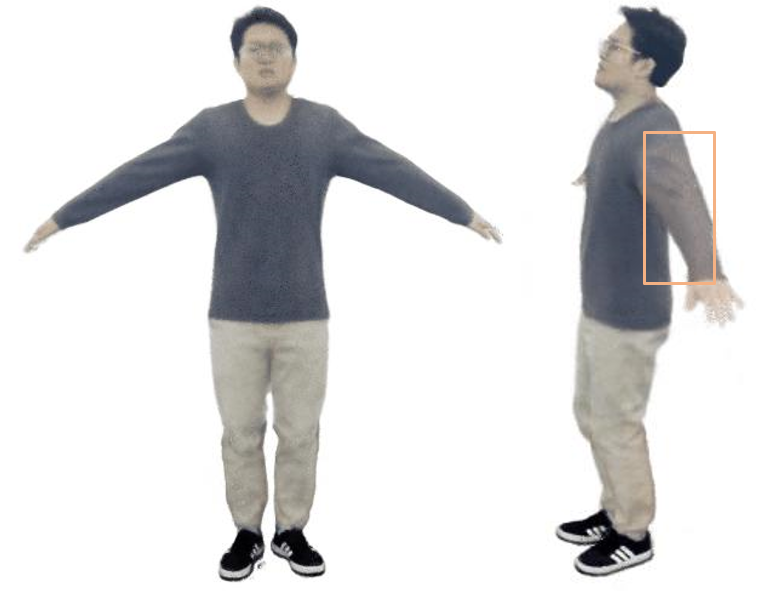}
	\end{minipage}
	}   
	\subfigure[w/o viewing direction]{
	\begin{minipage}[b]{0.33\linewidth}
	\includegraphics[width=1.0\linewidth, trim=0 0 0 0,clip]{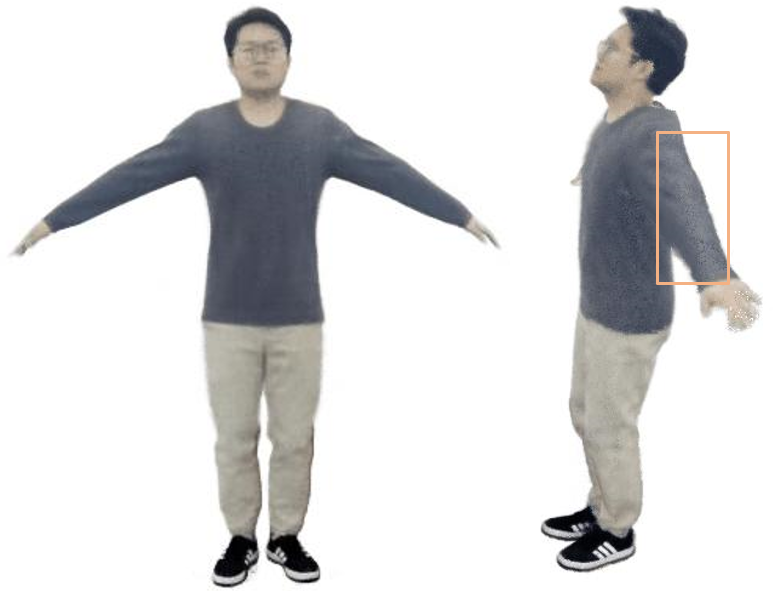}
	\end{minipage}
	} 
    \caption{Impact of viewing direction on novel view synthesis on iPER. After removing viewing direction from the input, our model produces more consistent result across different views.}
	\label{Impact_of_view_direction}
\end{figure}

\begin{figure}[ht]
    \centering
    \subfigure[Input]{
	\includegraphics[width=0.20\linewidth]{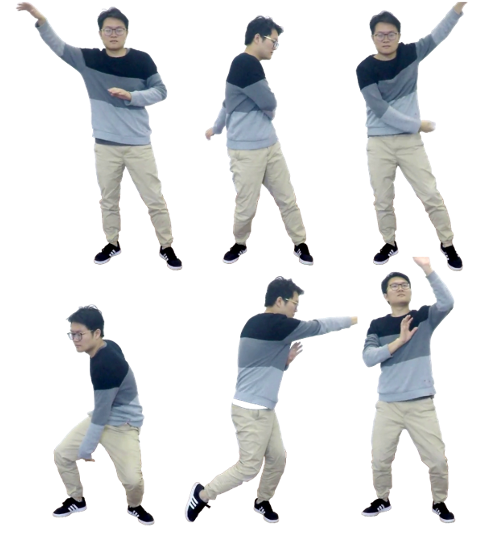}
	}
	\subfigure[w/o pose refinement]{
	\includegraphics[width=0.34\linewidth]{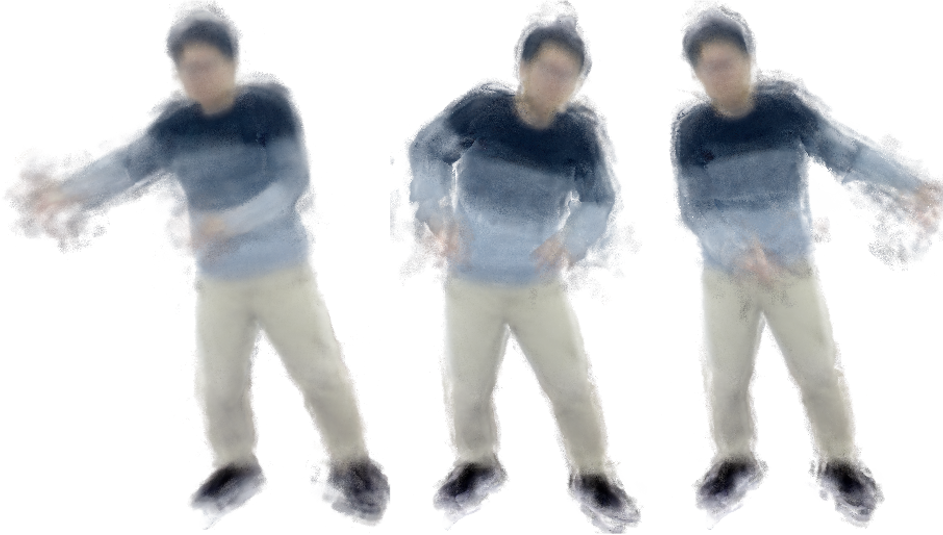}
	}
	\subfigure[Ours]{
	\includegraphics[width=0.35\linewidth]{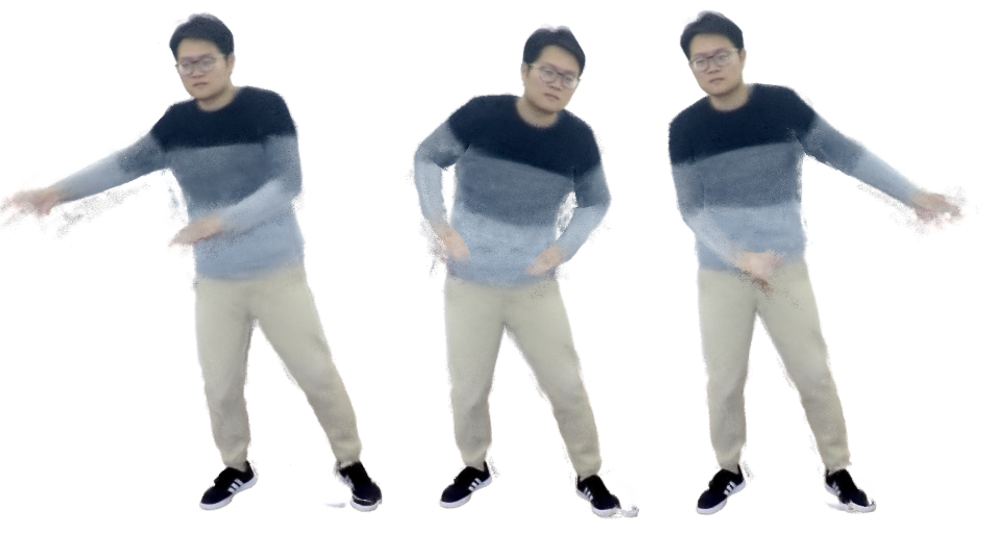}
	}
    \caption{Visualization result of novel pose synthesis on a complex pose video 009-4-2 of the iPER\cite{LWGAN}.}
    \label{complex_pose_result}
\end{figure}

\subsection{Impact of viewing direction}
\label{impact_of_view_direction}

Unlike NeRF\cite{NeRF}, which maps the 3D position and viewing direction 
to color and density, our approach excludes viewing direction from our input for robust dynamic reconstruction.
NeRF's viewing direction is mainly used to handle specular reflections for materials such as glass and metal. 
For dynamic scenes, it is very difficult to deal with changes of illumination, so we assume that the appearance of the subject is not view-dependent in our experiment.
Also, human skin and clothes are mainly diffuse reflective materials, and there exist very few specular reflections.
As shown in the Fig.~\ref{Impact_of_view_direction}, the novel view synthesis results generated with viewing direction as a condition during training show unpredictable artifacts.

\section{Limitations}

Our method reconstructs a detailed 3D human body model and renders realistic images from a monocular video. 
Typically, the training videos capture subjects turning around before the camera and holding an A-pose or T-pose.
When trained on a video containing complex poses, our method still obtains reasonable results (Fig. \ref{complex_pose_result}). But noticeable losses of details are observed compared to previous simple training videos.
The main reason is that it is more challenging to obtain accurate enough SMPL estimations for videos containing complex poses.
%
Another limitation is that it is difficult for our method to handle extremely loose clothes or complex non-rigid deformations of the garments, 
because our explicit pose-guided deformation associates spatial points to SMPL mesh, without explicit modeling of the garments.
%
Thus, the novel view synthesis results inevitably lose some details on the clothes. So, to the get best results, the performer should slowly turn around and hold a simple pose so that their clothes almost remain still relative to their body for high-quality rendering. 
%
%
Like all NeRF based methods trained for only one scene, Our method cannot reconstruct invisible parts, such as the underarms and the inner thighs, so the input video needs to cover the whole body of the human body as much as possible. 

\section{Conclusion}
In this paper, we propose to learn an animatable neural radiance field from a monocular video, which allows us to perform photo-realistic novel-view synthesis, reconstruct the 3D geometry of the person with high-quality details, and animate the person with novel poses. 
To achieve these goals, we extend the neural radiance field to dynamic scenes with human movements via introducing an explicit pose-guided deformation module and an analysis-by-synthesis pose refinement strategy. 
Specifically, the pose-guided deformation attempts to deform the 3d position according to the neighboring SMPL vertices to learn a good and controllable human template in the canonical space, as well as to learn accurate 3d geometry.
The pose refinement strategy compensates for the negative impact of inaccurate pose estimation from existing approaches and provides more consistent
guidance for learning better geometry (i.e. density) and appearance (i.e. RGB). 
Experiments on both synthetic data and real data demonstrate the effectiveness of the proposed approach. 
%


\bibliographystyle{IEEEtran}
\bibliography{reference}

\end{document}